\documentclass{article} 

\PassOptionsToPackage{numbers}{natbib}

    \usepackage[preprint]{include/neurips_2019}

\usepackage[utf8]{inputenc} 
\usepackage[T1]{fontenc}    
\usepackage{hyperref}       
\usepackage{url}            
\usepackage{booktabs}       
\usepackage{amsfonts}       
\usepackage{nicefrac}       
\usepackage{xfrac}       
\usepackage{microtype}      

\usepackage{pgfplotstable}
\usepackage{tikz}
\usepackage{pgfplots}
\usepackage{collcell}
\usepackage{xstring}
\usepackage{tabularx}
\usepackage{graphicx}
\usepackage{tabu}
\usepackage{environ}
\usepackage{fp}
\usepackage{upgreek}
\usepackage{amsmath}
\usepackage{amsthm}
\usepackage{algorithm, algorithmic}
\usepackage{enumitem}
\usepackage{bm}
\usepackage{subfiles}
\usepackage{array,multirow}
\usepackage{caption,subcaption}
\usepackage{makecell}

\title{Generalizing from a few environments in safety-critical reinforcement learning}

\author{Zachary Kenton$^{1}$ \
Angelos Filos$^1$ \
Owain Evans$^2$\
Yarin Gal$^1$ \
\\
$^1$Oxford Applied \& Theoretical Machine Learning, University of Oxford \\
$^2$Future of Humanity Institute, University of Oxford
\\
\texttt{zachary.kenton@cs.ox.ac.uk}
}

\begin{document}

\maketitle

\begin{abstract}
Before deploying autonomous agents in the real world, we need to be confident they will perform safely in novel situations. Ideally, we would expose agents to a very wide range of situations during training, allowing them to learn about every possible danger, but this is often impractical. This paper investigates safety and generalization from a limited number of training environments in deep reinforcement learning (RL). We find RL algorithms can fail dangerously on unseen test environments even when performing perfectly on training environments. Firstly, in a gridworld setting, we show that catastrophes can be significantly reduced with simple modifications, including ensemble model averaging and the use of a blocking classifier. In the more challenging CoinRun environment we find similar methods do not significantly reduce catastrophes. However, we do find that the uncertainty information from the ensemble is useful for predicting whether a catastrophe will occur within a few steps and hence whether human intervention should be requested.
\end{abstract}

\section{Introduction} \label{sec:Introduction}

\paragraph{Problem Setting.}
Recent progress in deep reinforcement learning (RL) has achieved impressive results
in a range of applications from playing games \citep{mnih2015human, silver2016mastering},
to dialogue systems \citep{li2017end}
and robotics \citep{levine2016end,andrychowicz2018learning}.
However, generalizing to unseen environments remains difficult for deep RL algorithms, which can fail catastrophically
when encountering new environments \citep{leike2017ai}.
We consider the setting where an RL agent trains on a limited number of environments and must generalize to unseen environments. The agent will not perform perfectly on the unseen environments. But can it avoid dangers that were already encountered during training?

In safety-critical domains there can be catastrophic outcomes which are unacceptable -- see
\citep{garcia2015comprehensive} for a review on safety in RL. We would ideally like our RL agents to be able to avoid the dangers consistent with those seen during training, without requiring a hand-crafted safe policy for these.

In this work, we assume that we have access
to a simulator, which captures the basic semantics of the world
(i.e. dangers, goals and dynamics). 
In the simulator the agent can experience dangers and learn from them~\citep{paul2018fingerprint}.
We evaluate agents on how well they can transfer knowledge: can they generalize to unseen environments with the same basic semantics?
At deployment, the agent has a single episode to solve
an unseen environment and any dangerous behaviour is considered an unacceptable catastrophe. 

\paragraph{Related Work.}
Motivated by the standard regularization methods for tackling overfitting in deep
neural networks, \citet{farebrother2018generalization} and \citet{cobbe2018quantifying}
experiment with L2-regularisation, dropout \citep{srivastava2014dropout} and batch normalization
\citep{ioffe2015batch} with Deep Q-Networks \citep{mnih2015human}, showing improved
generalization performance.

\citet{zhang2018study} investigate the ability of A3C
\citep{mnih2016asynchronous} to generalize rather than memorize in a set
of gridworlds similar to our environments. They show that perfect generalization is
possible when a sufficient amount of environments is provided (10000 environments), but
they do not focus on the regime of a limited number of training environments, nor evaluate
performance in terms of safety.
Similarly, the focus of \citet{cobbe2018quantifying} is on a large number of training environments. 
At the other extreme, \citet{leike2017ai}
introduce a `Distribution Shift' gridworld setup, where they 
train on a single environment and deploy on another.

In a different direction, \citet{saunders2018trial} approached danger avoidance by using
supervised learning to train a blocker (i.e. a classifier) using a human-in-the-loop to maintain safety during training, which
restricts its scalability. 
A collision prediction model was also considered in the model-based setting in \citet{kahn2017uncertainty}.
In \citet{lipton2016combating}, catastrophes are avoided by training an intrinsic fear model to predict whether
a catastrophe will occur, and using this to perform reward shaping.

From a modeling perspective, an ensemble of models often performs better than a single model \citep{dietterich2000ensemble}. They can also be used for predictive uncertainty estimation of deep neural networks \citep{lakshminarayanan2017simple}. In our work we make use of this uncertainty estimation.

Finally, our approach can also be related to meta-learning \citep{schmidhuber1987evolutionary, thrun2012learning, hochreiter2001learning, bengio1992optimization}, which is concerned with learning strategies which are fast to adapt using prior experience. In the RL context, approaches include gradient-based \citep{finn2017model} and recurrent style \citep{wang2016learning, duan2016rl} models using multiple environments to train from. Our setting corresponds to the zero-shot meta-RL setting, in which we train on multiple training environments but do not adapt based on test environment reward signals.

\paragraph{Contributions.}
We first investigate safety and generalization in a class of gridworlds. We find that standard DQN fails to avoid catastrophes at test time, even with 1000 training environments. We compare standard DQN to modified versions that incorporate dropout, Q-network ensembling, and a classifier to recognize dangerous actions. These modifications reduce catastrophes significantly, including in the regime of very few training environments.  
We next look at safety and generalization in the more challenging CoinRun environment. We find that in this case simple model averaging does not help significantly to reduce catastrophes compared to a PPO baseline. 
However, we find that there is still important uncertainty information captured in the ensemble of value functions of the PPO agents. We perform a study on whether the agent can predict ahead of time whether a catastrophe will occur, given the information in the ensemble of value functions. We find that the uncertainty in these value functions is helpful for predicting a catastrophe. This is useful as it can be used to improve safety by requesting an intervention from a human.

\section{Background} \label{sec:Background}

\paragraph{Task Setup.}
We consider an agent interacting with an environment in the standard
RL framework \citep{sutton2018reinforcement}. At each step,
the agent selects an action based on its current state,
and the environment provides a reward and the
next state.
Our task setup is the same as in \citep{zhang2018study}: there is a train/test split for \textit{environments} that is analogous to the train/test split for \textit{data points} in supervised learning. In our experiments all environments will have the same reward and transition function, and differ only in the initial state. Hence we can equivalently describe our setup in terms of a distribution on initial states for a single  MDP.

Formally, we denote our task by
$(\mathcal{M}=(\mathcal{S}, \mathcal{A}, \mathcal{P}, R), \mathcal{P}_{0})$,
where $\mathcal{M}$ is a Markov Decision Process (MDP), with state space $\mathcal{S}$, action space
$\mathcal{A}$,
transition probability $\mathcal{P}$ and immediate
reward function $r$. 
Additionally, $\mathcal{P}_{0}$ is a probability distribution
on the initial state $\mathcal{S}_{0} \subset \mathcal{S}$.
We use the undiscounted episodic setting, where each episode randomly samples an initial state from $\mathcal{P}_{0}$ and ends in a finite number of timesteps, $T$. There are disjoint training and test sets which have i.i.d. samples $\hat{\mathcal{S}}_0$ from $\mathcal{P}_{0}$. During training the agent encounters initial states only from the training set and makes learning updates based on the observed rewards. Test performance is calculated on the test set, and no learning takes place at test time.

\section{Gridworld Experiments} \label{sec:Experiments}

\subsection{Experimental Setup}
\begin{figure}[ht]
  \centering
  \begin{subfigure}[l]{0.15\linewidth}
    \includegraphics[width=\linewidth]{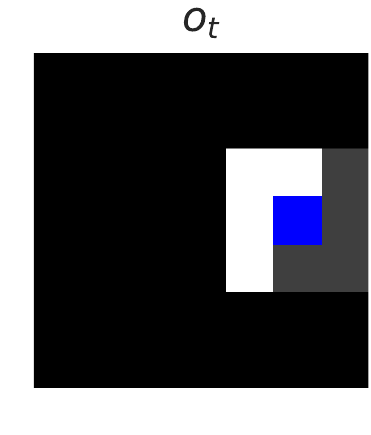}
  \end{subfigure}
  ~
  \begin{subfigure}[l]{0.15\linewidth}
    \includegraphics[width=\linewidth]{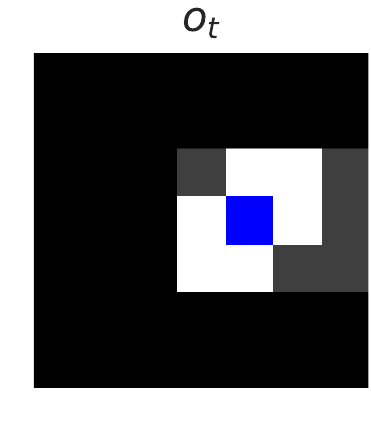}
  \end{subfigure}
  ~
  \begin{subfigure}[l]{0.15\linewidth}
    \includegraphics[width=\linewidth]{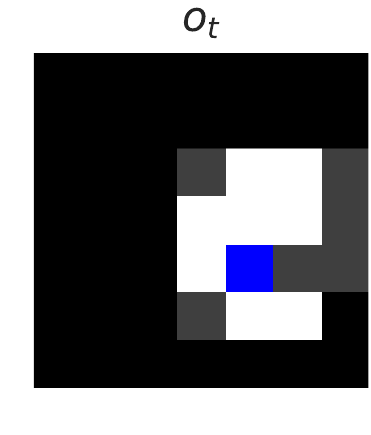}
  \end{subfigure}
  ~
  \begin{subfigure}[l]{0.15\linewidth}
    \includegraphics[width=\linewidth]{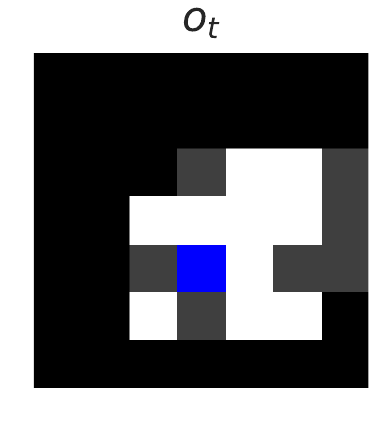}
  \end{subfigure}
  ~
  \begin{subfigure}[l]{0.15\linewidth}
    \includegraphics[width=\linewidth]{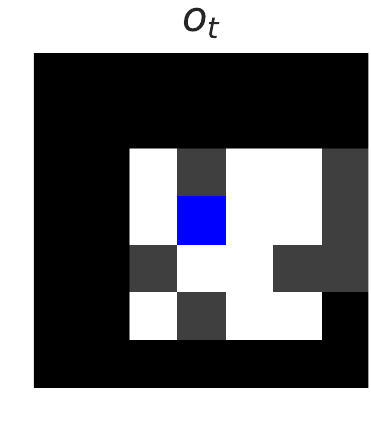}
  \end{subfigure}
  ~
  \begin{subfigure}[l]{0.15\linewidth}
    \includegraphics[width=\linewidth]{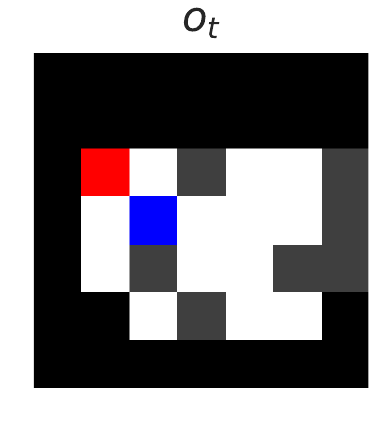}
  \end{subfigure}
  ~
  \begin{subfigure}[l]{0.15\linewidth}
    \includegraphics[width=\linewidth]{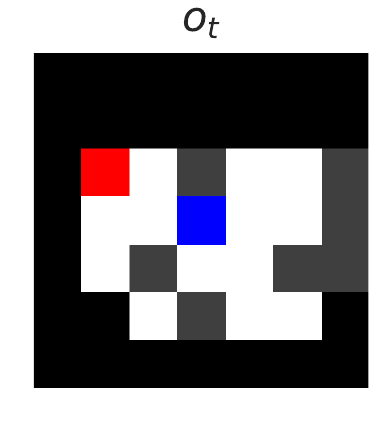}
  \end{subfigure}
  ~
  \begin{subfigure}[l]{0.15\linewidth}
    \includegraphics[width=\linewidth]{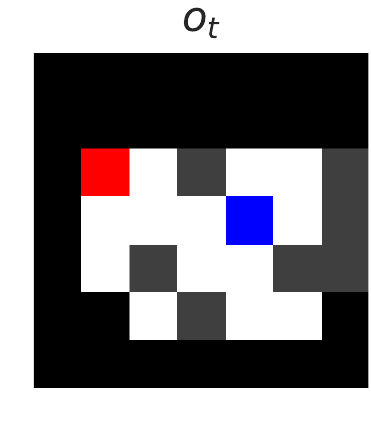}
  \end{subfigure}
  ~
  \begin{subfigure}[l]{0.15\linewidth}
    \includegraphics[width=\linewidth]{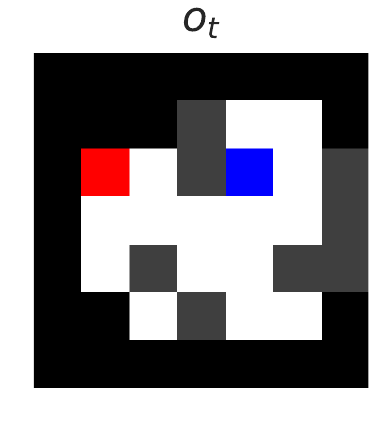}
  \end{subfigure}
  ~
  \begin{subfigure}[l]{0.15\linewidth}
    \includegraphics[width=\linewidth]{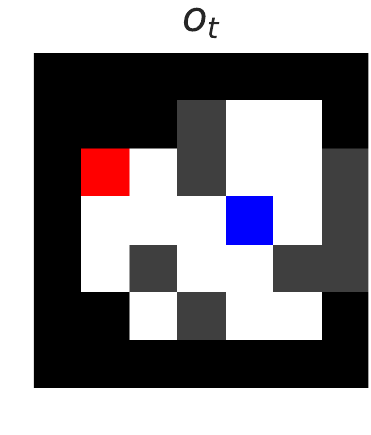}
  \end{subfigure}
  ~
  \begin{subfigure}[l]{0.15\linewidth}
    \includegraphics[width=\linewidth]{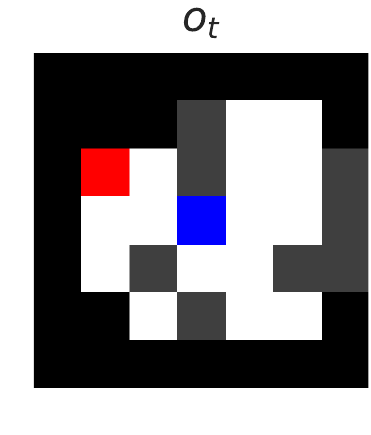}
  \end{subfigure}
  ~
  \begin{subfigure}[l]{0.15\linewidth}
    \includegraphics[width=\linewidth]{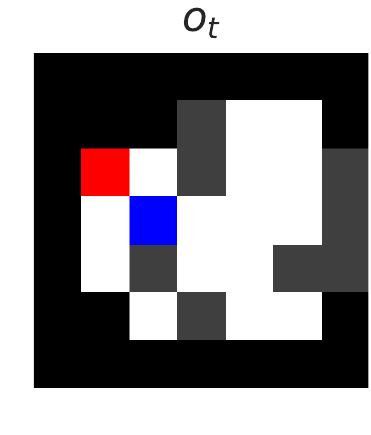}
  \end{subfigure}
  ~
  \begin{subfigure}[l]{0.15\linewidth}
    \includegraphics[width=\linewidth]{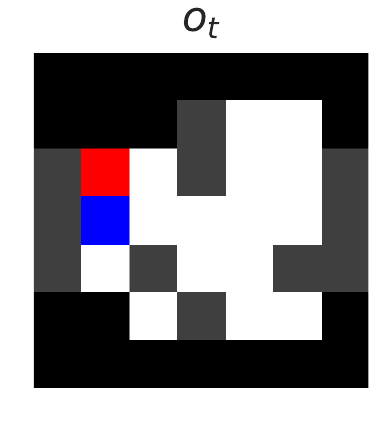}
  \end{subfigure}
  ~
  \begin{subfigure}[l]{0.15\linewidth}
    \includegraphics[width=\linewidth]{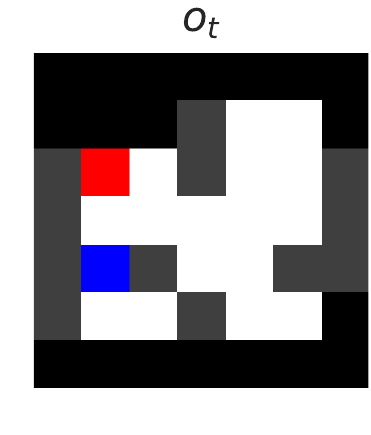}
  \end{subfigure}
  ~
  \begin{subfigure}[l]{0.15\linewidth}
    \includegraphics[width=\linewidth]{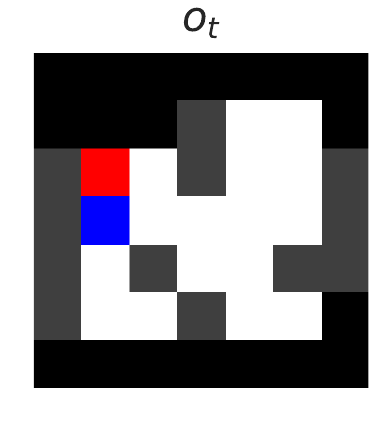}
  \end{subfigure}
  ~
  \begin{subfigure}[l]{0.15\linewidth}
    \includegraphics[width=\linewidth]{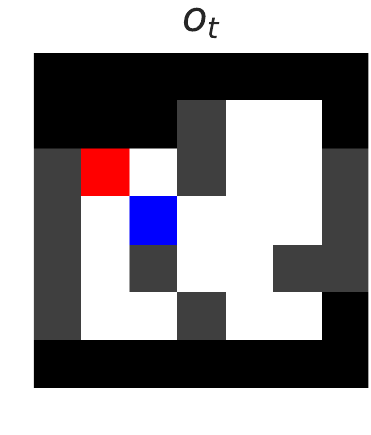}
  \end{subfigure}
  \begin{subfigure}[l]{0.15\linewidth}
    \includegraphics[width=\linewidth]{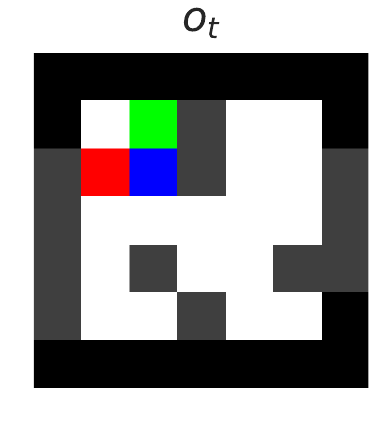}
  \end{subfigure}
  \begin{subfigure}[l]{0.15\linewidth}
    \includegraphics[width=\linewidth]{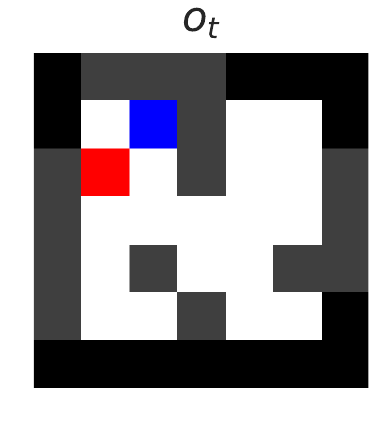}
  \end{subfigure}
  \caption{Example trajectory from a \texttt{Reveal} environment. Agent: blue. Goal: green. Lava: red. Walls: grey. Mask: black.}
  \label{fig:partial_obs}
\end{figure}

Our environment setup is a distribution of gridworld environments, each of which is size $5\times5$, and contains an agent (blue), a single lava cell (red) and a single goal cell (green). The agent receives sparse rewards of $+1$ for reaching the goal and $-1$ for reaching the lava. The episode terminates whenever the goal or lava is reached, or when fifty timesteps have elapsed (giving zero reward), whichever occurs first.
We consider two environment settings, which we call \texttt{Full} and \texttt{Reveal}. In \texttt{Full}, the agent sees the full map (an example trajectory is shown in Supplementary Material, Fig.~\ref{fig:full_obs}), whereas in \texttt{Reveal}, Fig.~\ref{fig:partial_obs}, the agent starts off seeing only part of the map, and reveals the map as it goes around, with a $3\times3$ view. \texttt{Reveal} is a more challenging setting because it requires the agent to move around to uncover the position of the goal. The agent receives the observation as an array of RGB pixel values flattened across the channel dimension.

We treat moving onto the lava as a catastrophe. Our evaluation metrics are the percentage of environments that are solved (the agent reaches the goal before the timeout), and the percentage of environments that end in catastrophe (the agent reaches the lava). On test environments we consider timeouts to be an acceptable failure, whereas a catastrophe is unacceptable.

\subsection{Methods}
\paragraph{Deep Q-Networks (DQN).}
Deep Q-networks \citep{mnih2015human} do Q-learning \citep{watkins1992q} using a deep neural network as a function approximator to estimate the optimal value function $Q(s,a; \theta)$, where $\theta$ is a parameter vector. 
DQN is optimized by minimizing
$L_i(\theta_i) = \mathbb{E}_{s,a,r,s'}[(y_i - Q(s,a; \theta_i))^2]$, at each iteration $i$, where $y_i=r+\max_{a'}Q(s', a'; \theta^-)$. The $\theta^-$ are parameters of a target network that is kept frozen for a number of iterations whilst updating the online network parameters $\theta$.
The optimization is performed off-policy, randomly sampling from an experience replay buffer. During training, actions are chosen using the $\epsilon$-greedy exploration strategy, selecting a random action with probability $\epsilon$ and otherwise taking the greedy action (which has maximum Q-value). At test time, the agent acts greedily.

\paragraph{Model Averaging.}
Ensembles of models (i.e. \textit{model averaging}) are usually used for
estimating model (i.e. \textit{epistemic}) uncertainty. In particular,
instead of a single model $f$, a set of models $f_{1}, f_{2}, \ldots, f_{N}$
is fitted. Then either the average,
$f_{\text{ens}} = \frac{1}{N} \sum_{n=1}^{N} f_{n}$
or, in classification tasks, the mode (i.e. \textit{majority vote})
$f_{\text{maj}} = \mathtt{mode}(f_{1}, f_{2}, \ldots, f_{N})$
is used for prediction. When neural networks are used as models,
the diversification between the models is obtained by initializing
them differently and by following independent training. 
For model averaging on DQN, we do the model averaging on the $Q$-value.

\paragraph{Catastrophe Classifier.}
Another approach to avoiding dangers is to learn a classifier for whether a state-action pair will be catastrophic and use this to block certain actions --- see \citep{saunders2018trial} for an example trained with a human-in-the-loop. 
During training we store all state-action pairs, together with a binary label of whether a catastrophe occurred. Then after training the DQN agent, we separately train the classifier to predict the probability that a state-action pair will result in a catastrophe. Training is done in a supervised manner by minimizing the binary cross entropy loss. The classifier is used as a `blocker' at deployment time. At test time we run our selected action through the classifier and
block the action if the classifier predicts it is catastrophic with confidence greater than some threshold. We then move on to the next highest value action and run that through the classifier. The process repeats until an acceptable action is found, otherwise the episode is terminated.
Note that the blocker will only block dangerous actions that occur just before the danger is about to be experienced, but won't help for those actions which irreversibly cause a catastrophe to occur many steps later \citep{saunders2018trial}.

\paragraph{Algorithm Settings}
A summary of the methods used can be found in Tab.~\ref{tab:settings_hparams}. All 3-layer multi-layer
perceptron DQN models were trained for 1M training episodes
using: hidden layer sizes [256,256,512], batch size 32, RMSProp \citep{tieleman2012lecture} with learning rate 1e--4,
a replay buffer with 10K capacity and the target network was updated every 1K
episodes. An $\epsilon$-greedy policy was used with an exponential decay rate 0.999
and end value 0.05. The blocker is also a 3-layer
multi-layer perceptron with hidden layer sizes [128,256,256] trained for 10k iterations using: batch size 64,
Adam optimizer \citep{kingma2014adam} with learning rate 5e--3.

\begin{table*}[!bt]
\newcommand{\factor}{}
\small
\centering
\begin{tabular}{ll}
\toprule
\bf{Method} & \bf{Description} \\
\cmidrule(r){1-1} \cmidrule(lr){2-2}
DQN & Same as \citep{mnih2015human} \\
Drop-DQN & Regularized linear layers with dropout probability $p=0.2$ \\
Block-DQN & Catastrophe classifier used along with DQN \\
Ens-DQN & Ensemble of 9 independently trained and differently intialized DQNs \\
Maj-DQN & Majority vote of 9 independently trained and differently intialized DQNs \\
\midrule
Block\&Ens-DQN & Combination of Block-DQN and Ens-DQN \\ 
\bottomrule
\end{tabular}
\caption{Description of methods used in our Gridworld experiments. }
\label{tab:settings_hparams}
\end{table*}

\subsection{Results and Discussion.}
\begin{figure}[ht]
  \begin{subfigure}[b]{\linewidth}
    \includegraphics[width=\linewidth]{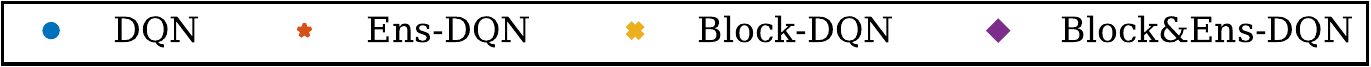}
  \end{subfigure}

  \begin{subfigure}[t]{.48\linewidth}
    \includegraphics[width=\linewidth]{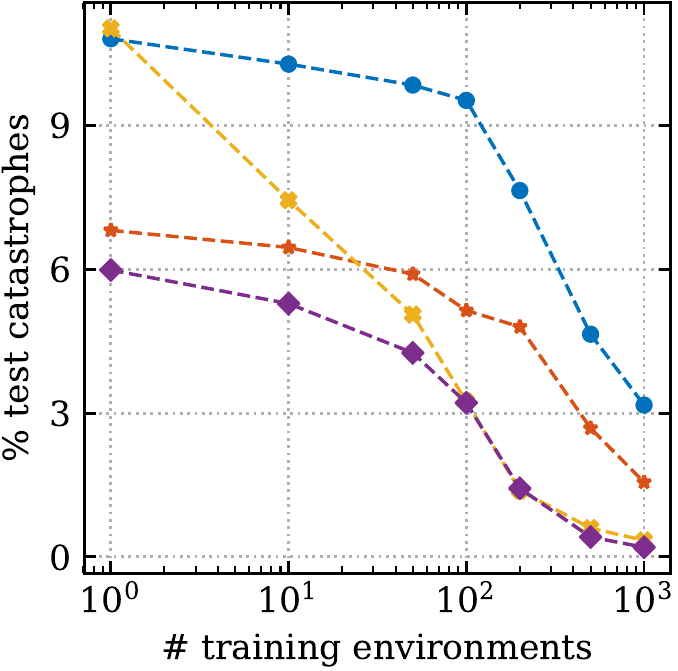}
    \caption{Percentage of catastrophic outcomes in unseen environments (lower is better),
    as a function of number of training environments.}
    \label{fig:reveal::catastrophes}
  \end{subfigure}
  ~
  \begin{subfigure}[t]{.48\linewidth}
    \includegraphics[width=\linewidth]{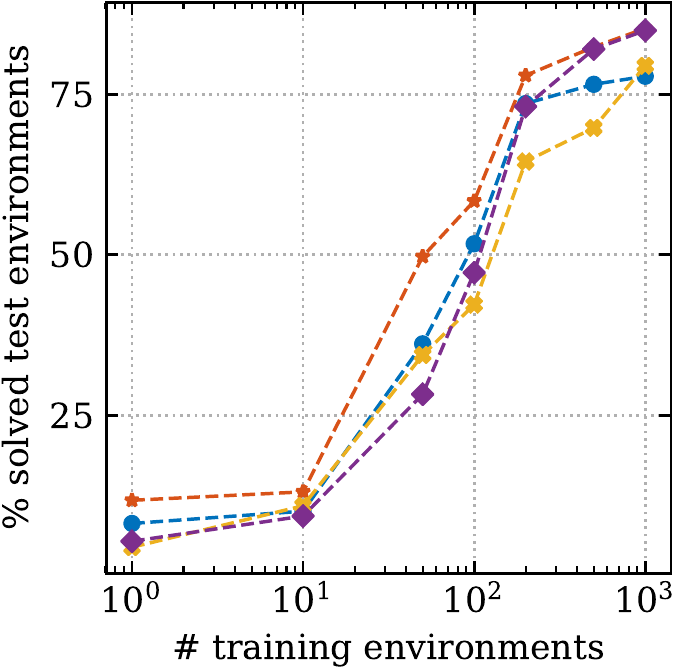}
    \caption{Percentage of solved unseen environments (higher is better),
    as a function of training environments.}
    \label{fig:reveal::solved}
  \end{subfigure}

  \caption{Results on the \texttt{Reveal} setting, 
  evaluated on unseen \textit{test} environments for
  a range of methods. Nine random seeds are used for each algorithm and mean performances is shown here. Figure \textbf{(a)} shows that modified algorithms
  outperform the baseline DQN in terms of danger avoidance.
  The effect on return performance is observed in \textbf{(b)}.
  The complete version is provided in Figure 
  \ref{fig:reveal::quant_results_complete} of the appendix, and includes both train and test performances.}
  \label{fig:reveal::quant_results}
\end{figure}

To make figures easier to read, this section includes only four methods: DQN,
Ens-DQN, Block-DQN and Block\&Ens-DQN.
In Fig.~\ref{fig:reveal::quant_results} we present results on the \texttt{Reveal} gridworld. 
We plot the  percentage of environments that ended in catastrophe in Fig.~\ref{fig:reveal::catastrophes}, and the percentage of solved environments in Fig.~\ref{fig:reveal::solved}, as a function of the number of training environments available during training.
We trained all models to convergence on the training environments.
See Fig.~\ref{fig:full::quant_results_complete} and Fig.~\ref{fig:reveal::quant_results_complete} in supplementary material for results of all methods on \texttt{Full} and \texttt{Reveal} settings and also for the evaluations on the training environments.

Fig.~\ref{fig:reveal::solved} shows that our agents never achieve perfect performance on the test environments. Moreover, when an agent fails to reach the goal, it does not always fail gracefully (e.g.\ by simply timing out) but instead often ends in catastrophe (visiting the lava). 

Most of the methods we investigated outperformed the DQN baseline in terms of percentage of test catastrophes. Each method offers a different trade-off between test performance on catastrophes and solved environments. For example, Block-DQN offers better catastrophe performance than DQN, but its performance on solving environments is worse given more than 100 training environments. 
This is possibly because the blocker is over-cautious, with too high a false-positive rate for catastrophes, which prematurely stops some environments from being solved. Note that in a real-world setting, avoiding catastrophes (Fig.~\ref{fig:reveal::catastrophes}) will be much more important than doing well on most environments (Fig.~\ref{fig:reveal::solved}). 

\begin{figure}[!htbp]
  \begin{subfigure}[b]{0.35\linewidth}
    \centering
    \includegraphics[width=0.6\linewidth]{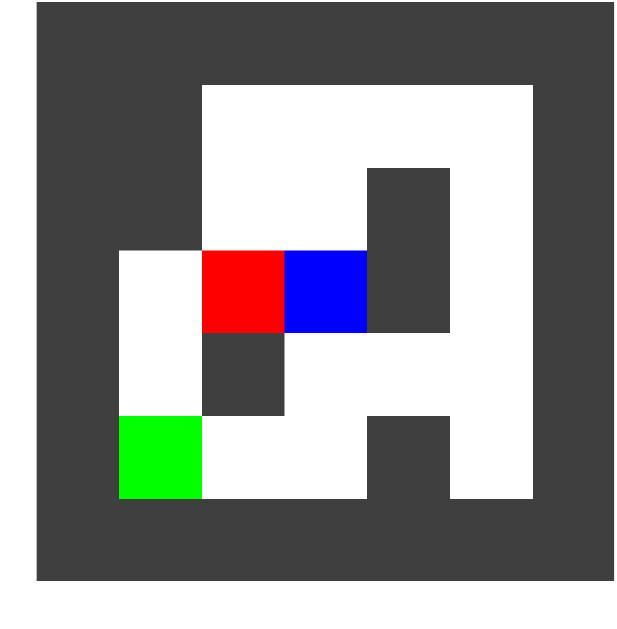} \\
    \caption{$s_{t}$} \vspace{0.25cm}
    \includegraphics[width=0.6\linewidth]{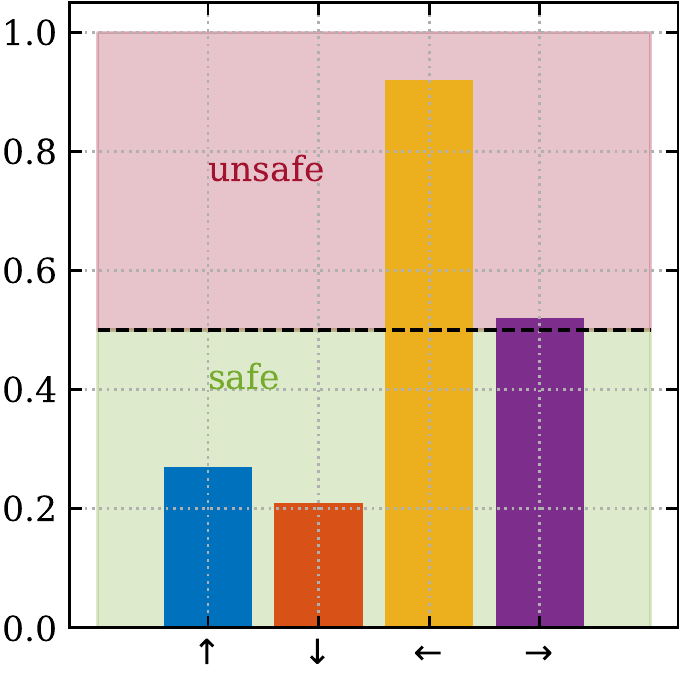}
    \caption{$p_{\text{unsafe}}(a_{t} | s_{t})$}
    \label{subfig:combo}
  \end{subfigure}
  ~
  \begin{subfigure}[b]{0.45\linewidth}
    \includegraphics[width=1.0\linewidth]{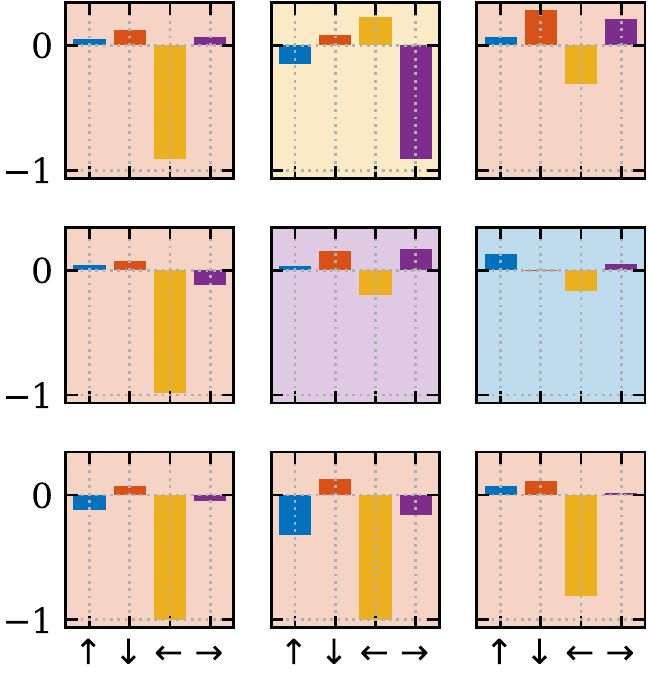}
    \caption{$Q^{(i)}(s_{t}, a_{t}),\ \text{for}\ i=1,2,\ldots,9$}
    \label{subfig:Ens-DQN}
  \end{subfigure}
  \caption{Example transition by the Block\&Ens-DQN in one unseen environment,
  in the \texttt{Full} setting. \textbf{(a)} the environment state, $s_{t}$;
  \textbf{(b)} the output of the trained catastrophe classifier (i.e. \textit{blocker})
  $p_{\text{unsafe}}(\cdot | s_{t})$ conditioned on the environment state,
  where a threshold 0.5 is selected; \textbf{(c)} the nine estimates of the state-action
  value function $Q^{(i)}(s_{t}, a_{t}),\ \text{for}\ i=1,2,\cdots,9$, from the differently
  initialized and independently trained DQNs. The background colour highlights action
  with maximum value. The agent should not make the catastrophic
  action of going \textit{left}, something that both the blocker and
  the ensemble (i.e. \textit{model average}) of the DQNs will avoid.
  However, if the middle top agent in \textbf{(c)} was acting
  alone it would choose to go left, which would lead to a catastrophic outcome.}
  \label{fig:failure_cases}
\end{figure}

In Fig.~\ref{fig:failure_cases} we showcase an example state from our experiments highlighting the role of the ensemble and the blocker in avoiding the catastrophe. 

\section{CoinRun Experiments}

\subsection{Experimental Setup}
\begin{figure}
    \begin{subfigure}[b]{0.45\linewidth}
    \centering
         \includegraphics[width=\columnwidth]{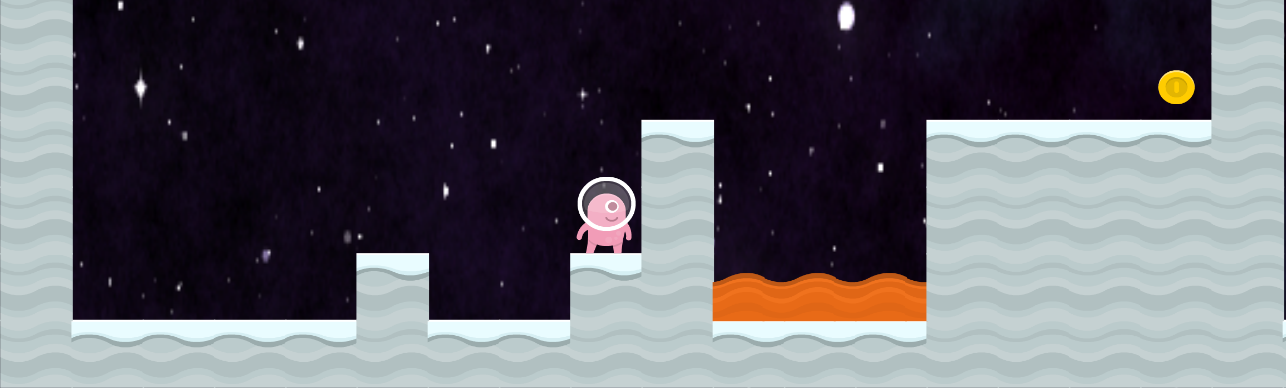}
  \end{subfigure}
  \begin{subfigure}[b]{0.5\linewidth}
    \centering
         \includegraphics[width=\columnwidth]{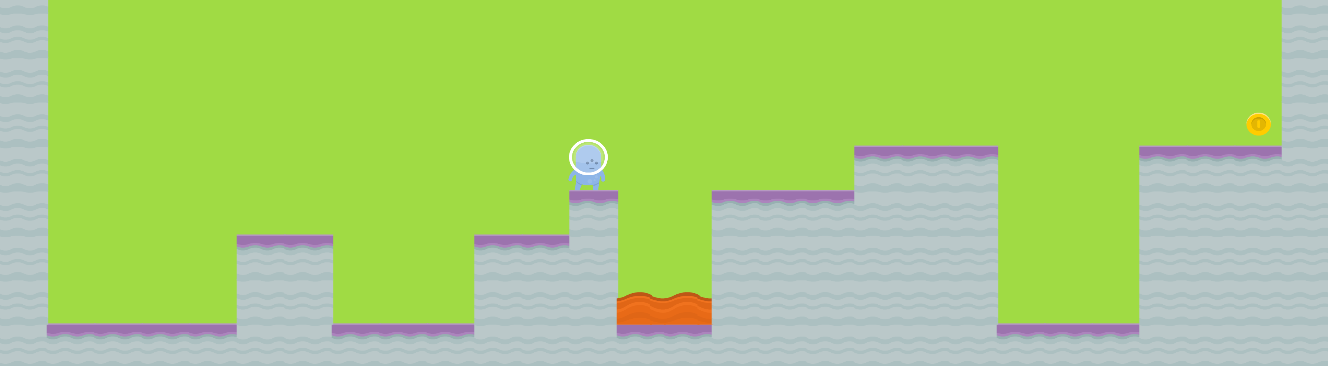}
  \end{subfigure}
    \caption{Two sample environments from our modified CoinRun setting.}
    \label{fig:coinrun:screenshot}
\end{figure}

Following our experiments on gridworlds, we next consider the more challenging CoinRun environment \citep{cobbe2018quantifying}, a procedurally generated game in which the agent is spawned on the left and whose aim is to reach the coin on the right whilst avoiding obstacles, see Fig.~\ref{fig:coinrun:screenshot} for some screenshots. The agent receives a reward of +5 for reaching the coin, and the episode terminates with -5 reward either after 1000 timesteps, or on collision with an obstacle.
We simplified the environment from \citep{cobbe2018quantifying} to remove all crates and obstacles except for the lava and to have only six actions (no-op, jump, jump-right, jump-left, right, left). 
This simplification allowed us to train our agents in 10 million timesteps, rather than 256 million. 
In our setup we consider falling in the lava to be a catastrophe, whereas a timeout is an acceptable failure. The observations given to the agent is the RGB 64x64 pixel values, flattened along the channel dimension.

\subsection{Methods}
\paragraph{Proximal Policy Optimization (PPO)}
In these experiments we use the Proximal Policy Optimization (PPO) algorithm \citep{schulman2017proximal} as it was shown by \citet{cobbe2018quantifying} to perform fairly well in the original CoinRun environment. We train five PPO agents independently and with different random initializations on each of 10, 25, 50 and 200 training levels. We used model averaging using majority vote (mode of sampled actions from the five agents), denoted Maj-PPO, and the sampling from the ensemble mean, denoted Ens-mean (where the mean distribution is formed by taking the mean over the logits of the individual PPO policy categorical distributions). We also trained a single agent with dropout for each of the  10, 25, 50 and 200 training levels. For full algorithm settings see Sec.~\ref{sec:app:algo} in the supplementary material.

\subsection{Results}
\begin{figure}[ht]
    
  \begin{subfigure}[b]{\linewidth}
    \centering
        \includegraphics[width=0.7\linewidth]{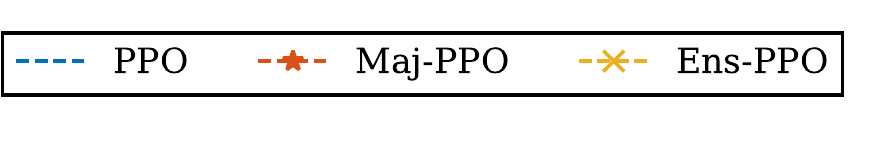}
  \end{subfigure}

  \begin{subfigure}[t]{0.48\linewidth}
  \centering
    \includegraphics[width=\linewidth]{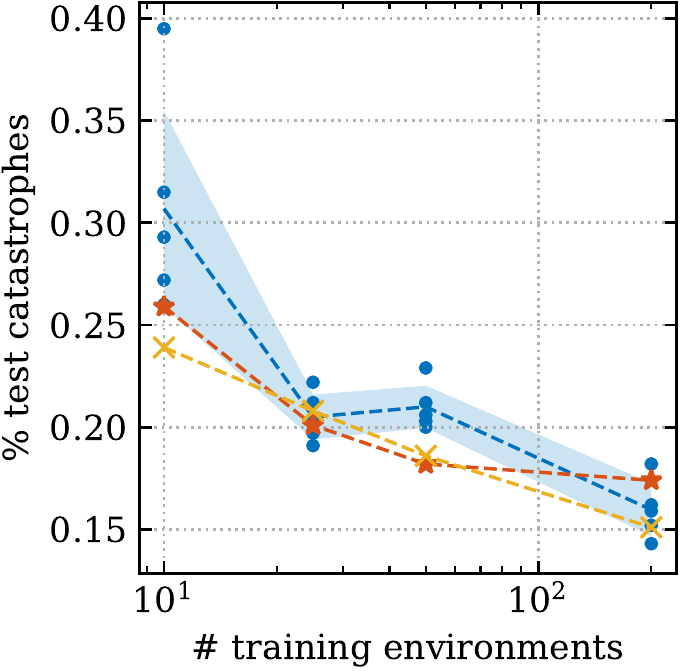}
    \caption{Percentage of catastrophic outcomes in unseen environments (lower is better),
    as a function of number of training environments.}
    \label{fig:coinrun::catastrophes}
  \end{subfigure}
  ~
  \begin{subfigure}[t]{0.48\linewidth}
  \centering
    \includegraphics[width=\linewidth]{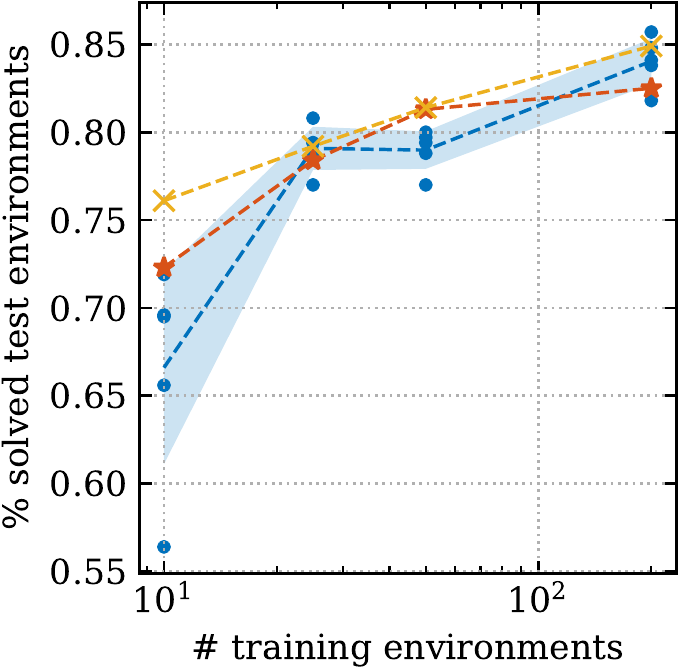}
    \caption{Percentage of solved unseen environments (higher is better),
    as a function of training environments.}
    \label{fig:coinrun::solved}
  \end{subfigure}

  \caption{Results in the CoinRun setting, 
  evaluated on unseen \textit{test} environments for
  a range of methods. Five random seeds are used for each algorithm. For the PPO baseline the dots mark the five seeds' performance, and the line and shading are the mean and one standard deviation intervals respectively. Other methods used all five seeds so no intervals appear for them. The ensemble algorithms don't do significantly better than a single PPO agent (on average) both in terms of catastrophes and \% solved.
  The complete version is provided in Fig.~ 
  \ref{fig:coinrun::quant_results_complete} of the appendix, and includes both train and test performances as well as dropout experiments.}
  \label{fig:coinrun::quant_results}
\end{figure}

\paragraph{Generalization Performance}
We plot the percentage of test levels ending in catastrophe and the percentage solved against the number of training environments in Fig.~\ref{fig:coinrun::quant_results}.  We see the two methods using an ensemble, Maj-PPO and Ens-mean, give similar performance to the baseline. We see a slight improvement for the ensembles for the 10 training environments setting. The other methods using dropout as a regularizer and MC dropout \cite{gal2016dropout} for ensembling did not match baseline performance, see Fig.~\ref{fig:coinrun::quant_results_complete} of the supplementary material, which also contains performance on the training set. We emphasise that performing perfectly on a small number of training environments is not sufficient to get good test performance, both for \% solved and more importantly for \% catastrophes. 

\paragraph{Predicting a Catastrophes in CoinRun}
 
In gridworld, the catastrophes are \emph{local} in that they occur exactly one step after the dangerous action is taken. In CoinRun catastrophes are \emph{non-local}: an agent takes a jump action and falls in the lava a few steps later (with no way to avoid the lava once in mid-air). We suspect this explains why it's harder to reduce catastrophes in CoinRun than gridworld.

Rather than modifying the agents actions, we instead now consider a setup where the agent should call for help if it thinks it has taken a dangerous action that will lead to a catastrophe. This is for example the intervention setup used in an autonomous driving application \cite{michelmore2018evaluating}.

The agent requests an intervention based on a discrimination function $U = \alpha \mu + \beta \sigma$ which combines the mean, $\mu$, and standard deviation, $\sigma$, of the ensemble of the five agents' value functions, similar to UCB \citet{Auer:2003:UCB:944919.944941}. 
We consider a binary classification task with a catastrophe occurring in $dt$ timesteps as the `positive` class, and predicting no catastrophe occurring in $dt$ timesteps as the `negative` class.
A true positive would be an agent predicting the catastrophe and it occurring, whereas a false positive would be predicting a catastrophe and it not occurring. We imagine a human intervention occurring on a positive prediction, and so would like to reduce the number of false positives (which might waste the human's time, or be suboptimal) and maintain a high true positive rate. 
An ROC curve captures the diagnostic ability of a binary classifier system as its discrimination threshold is varied -- the threshold is compared to the discrimination function $U$. In an ROC curve, the higher the sensitivity (true positive rate) and the lower the 1-specificity (false positive rate) the better. The AUC score is a summary statistic of the ROC curve, the higher the better.

\begin{figure}[ht]
    
  \begin{subfigure}[b]{\linewidth}
    \centering
        \includegraphics[width=\linewidth]{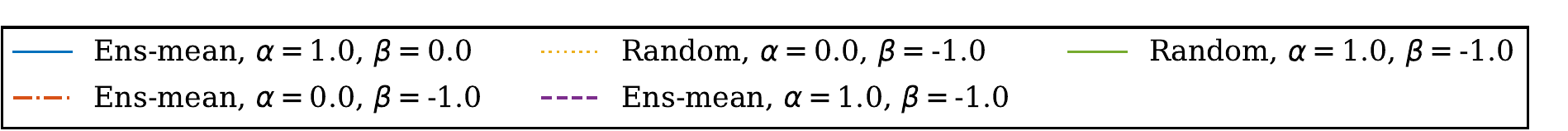}
  \end{subfigure}

  \begin{subfigure}[t]{0.48\linewidth}
  \centering
    \includegraphics[width=\linewidth]{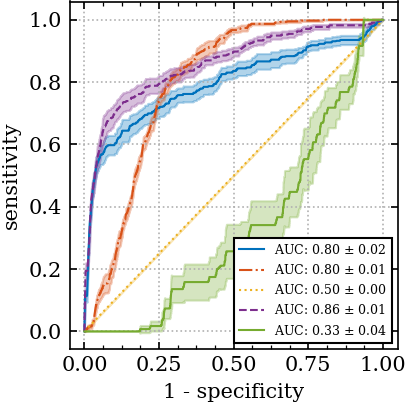}
    \caption{10 training levels, $dt=1$}
    \label{fig:roc:10t1}
  \end{subfigure}
  ~
  \begin{subfigure}[t]{0.48\linewidth}
  \centering
    \includegraphics[width=\linewidth]{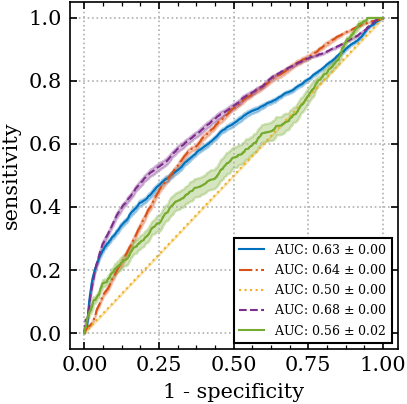}
    \caption{10 training levels, $dt=10$}
    \label{fig:roc:10t10}
  \end{subfigure}

  \vspace{1em}

  \begin{subfigure}[t]{0.48\linewidth}
  \centering
    \includegraphics[width=\linewidth]{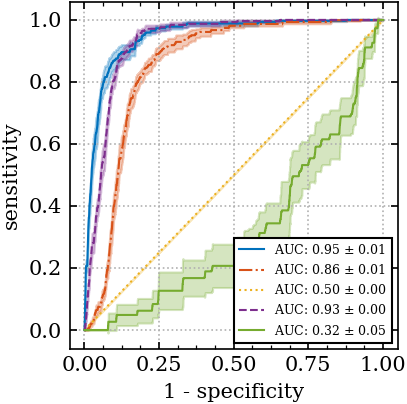}
    \caption{25 training levels, $dt=1$}
    \label{fig:roc:25t1}
  \end{subfigure}
    ~
  \begin{subfigure}[t]{0.48\linewidth}
  \centering
    \includegraphics[width=\linewidth]{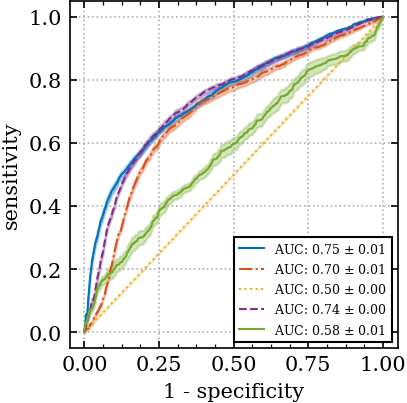}
    \caption{25 training levels, $dt=10$}
    \label{fig:roc:25t10}
  \end{subfigure}

  \caption{ROC Curves for binary classifier based on discrimination function $U = \alpha \mu + \beta \sigma$, a combination of mean and standard deviation of the agents' value functions, classifying a catastrophe occurring in the next $dt$ steps.  All plots are on data from one episode on 1000 different test levels. 
  Towards the top left is better.
  Higher AUC is better.
  \textbf{Left column:}  $dt=1$. 
  \textbf{Right column:} $dt=10$. 
  \textbf{Top row: } $10$ training levels. 
  \textbf{Bottom row:} $25$ training levels. }
  \label{fig:roc:main}
\end{figure}

In Fig~\ref{fig:roc:main} we plot ROC curves for the Ens-mean agent, together with an agent that has random value functions and takes random actions.  The different curves show different action selection methods (random or Ens-mean action selection), together with different discrimination function hyperparameters. Shown are mean and one standard deviation confidence intervals based on ten bootstrap samples from the data collected from one rollout on 1000 test environments.  

We see from Fig~\ref{fig:roc:10t1} that for 10 training levels and a prediction time window of one step, the uncertainty information from using the standard deviation of the ensemble-mean action selection method gives superior prediction performance compared to not using the standard deviation (shown also is that it's better than random). However, it helps less as the time-window increases, Fig~\ref{fig:roc:10t10}, or as the number of training levels increases, Fig~\ref{fig:roc:25t1}~\ref{fig:roc:25t10}. Note that it's much easier to predict a catastrophe for a smaller time window (left column), than a longer one (right column). This supports our hypothesis that it is the time-extended nature of the CoinRun danger which is particularly challenging to generalize about. 
See Fig.~\ref{fig:app_roc:test} and Fig.~\ref{fig:app_roc:train} in the supplementary material for a wider  range of ROC plots.

\section{Conclusion} \label{sec:Conclusion}

In this paper we investigated how safety performance generalizes when deployed on unseen test environments drawn from the same distribution of environments seen during training, where no further learning is allowed. We focused on the realistic case in which there are a limited number of training environments. We found RL algorithms can fail dangerously on the test environments even when performing perfectly during training. We investigated some simple ways to improve safety generalization performance. We also investigated whether a future catstrophe can be predicted in the challenging CoinRun environment, finding that uncertainty information in an ensemble of agents is helpful when only a small number of environments are available.

\clearpage

\bibliography{references}

\begin{thebibliography}{38}
\providecommand{\natexlab}[1]{#1}
\providecommand{\url}[1]{\texttt{#1}}
\expandafter\ifx\csname urlstyle\endcsname\relax
  \providecommand{\doi}[1]{doi: #1}\else
  \providecommand{\doi}{doi: \begingroup \urlstyle{rm}\Url}\fi

\bibitem[Abadi et~al.(2015)Abadi, Agarwal, Barham, Brevdo, Chen, Citro,
  Corrado, Davis, Dean, Devin, Ghemawat, Goodfellow, Harp, Irving, Isard, Jia,
  Jozefowicz, Kaiser, Kudlur, Levenberg, Man\'{e}, Monga, Moore, Murray, Olah,
  Schuster, Shlens, Steiner, Sutskever, Talwar, Tucker, Vanhoucke, Vasudevan,
  Vi\'{e}gas, Vinyals, Warden, Wattenberg, Wicke, Yu, and
  Zheng]{tensorflow2015-whitepaper}
Mart\'{\i}n Abadi, Ashish Agarwal, Paul Barham, Eugene Brevdo, Zhifeng Chen,
  Craig Citro, Greg~S. Corrado, Andy Davis, Jeffrey Dean, Matthieu Devin,
  Sanjay Ghemawat, Ian Goodfellow, Andrew Harp, Geoffrey Irving, Michael Isard,
  Yangqing Jia, Rafal Jozefowicz, Lukasz Kaiser, Manjunath Kudlur, Josh
  Levenberg, Dan Man\'{e}, Rajat Monga, Sherry Moore, Derek Murray, Chris Olah,
  Mike Schuster, Jonathon Shlens, Benoit Steiner, Ilya Sutskever, Kunal Talwar,
  Paul Tucker, Vincent Vanhoucke, Vijay Vasudevan, Fernanda Vi\'{e}gas, Oriol
  Vinyals, Pete Warden, Martin Wattenberg, Martin Wicke, Yuan Yu, and Xiaoqiang
  Zheng.
\newblock {TensorFlow}: Large-scale machine learning on heterogeneous systems,
  2015.
\newblock URL \url{http://tensorflow.org/}.
\newblock Software available from tensorflow.org.

\bibitem[Andrychowicz et~al.(2018)Andrychowicz, Baker, Chociej, Jozefowicz,
  McGrew, Pachocki, Petron, Plappert, Powell, Ray,
  et~al.]{andrychowicz2018learning}
Marcin Andrychowicz, Bowen Baker, Maciek Chociej, Rafal Jozefowicz, Bob McGrew,
  Jakub Pachocki, Arthur Petron, Matthias Plappert, Glenn Powell, Alex Ray,
  et~al.
\newblock Learning dexterous in-hand manipulation.
\newblock \emph{arXiv preprint arXiv:1808.00177}, 2018.

\bibitem[Auer(2003)]{Auer:2003:UCB:944919.944941}
Peter Auer.
\newblock Using confidence bounds for exploitation-exploration trade-offs.
\newblock \emph{J. Mach. Learn. Res.}, 3:\penalty0 397--422, March 2003.
\newblock ISSN 1532-4435.
\newblock URL \url{http://dl.acm.org/citation.cfm?id=944919.944941}.

\bibitem[Bengio et~al.(1992)Bengio, Bengio, Cloutier, and
  Gecsei]{bengio1992optimization}
Samy Bengio, Yoshua Bengio, Jocelyn Cloutier, and Jan Gecsei.
\newblock On the optimization of a synaptic learning rule.
\newblock In \emph{Preprints Conf. Optimality in Artificial and Biological
  Neural Networks}, pages 6--8. Univ. of Texas, 1992.

\bibitem[Cobbe et~al.(2018)Cobbe, Klimov, Hesse, Kim, and
  Schulman]{cobbe2018quantifying}
Karl Cobbe, Oleg Klimov, Chris Hesse, Taehoon Kim, and John Schulman.
\newblock Quantifying generalization in reinforcement learning.
\newblock \emph{arXiv preprint arXiv:1812.02341}, 2018.

\bibitem[Dietterich(2000)]{dietterich2000ensemble}
Thomas~G Dietterich.
\newblock Ensemble methods in machine learning.
\newblock In \emph{International workshop on multiple classifier systems},
  pages 1--15. Springer, 2000.

\bibitem[Duan et~al.(2016)Duan, Schulman, Chen, Bartlett, Sutskever, and
  Abbeel]{duan2016rl}
Yan Duan, John Schulman, Xi~Chen, Peter~L Bartlett, Ilya Sutskever, and Pieter
  Abbeel.
\newblock Rl$^2$: Fast reinforcement learning via slow reinforcement learning.
\newblock \emph{arXiv preprint arXiv:1611.02779}, 2016.

\bibitem[Espeholt et~al.(2018)Espeholt, Soyer, Munos, Simonyan, Mnih, Ward,
  Doron, Firoiu, Harley, Dunning, Legg, and Kavukcuoglu]{espeholt2018impala}
Lasse Espeholt, Hubert Soyer, Remi Munos, Karen Simonyan, Volodymir Mnih, Tom
  Ward, Yotam Doron, Vlad Firoiu, Tim Harley, Iain Dunning, Shane Legg, and
  Koray Kavukcuoglu.
\newblock Impala: Scalable distributed deep-rl with importance weighted
  actor-learner architectures, 2018.

\bibitem[Farebrother et~al.(2018)Farebrother, Machado, and
  Bowling]{farebrother2018generalization}
Jesse Farebrother, Marlos~C Machado, and Michael Bowling.
\newblock Generalization and regularization in dqn.
\newblock \emph{arXiv preprint arXiv:1810.00123}, 2018.

\bibitem[Finn et~al.(2017)Finn, Abbeel, and Levine]{finn2017model}
Chelsea Finn, Pieter Abbeel, and Sergey Levine.
\newblock Model-agnostic meta-learning for fast adaptation of deep networks.
\newblock In \emph{Proceedings of the 34th International Conference on Machine
  Learning-Volume 70}, pages 1126--1135. JMLR. org, 2017.

\bibitem[Gal and Ghahramani(2016)]{gal2016dropout}
Yarin Gal and Zoubin Ghahramani.
\newblock Dropout as a bayesian approximation: Representing model uncertainty
  in deep learning.
\newblock In \emph{international conference on machine learning}, pages
  1050--1059, 2016.

\bibitem[Garc{\i}a and Fern{\'a}ndez(2015)]{garcia2015comprehensive}
Javier Garc{\i}a and Fernando Fern{\'a}ndez.
\newblock A comprehensive survey on safe reinforcement learning.
\newblock \emph{Journal of Machine Learning Research}, 16\penalty0
  (1):\penalty0 1437--1480, 2015.

\bibitem[He et~al.(2016)He, Zhang, Ren, and Sun]{He_2016}
Kaiming He, Xiangyu Zhang, Shaoqing Ren, and Jian Sun.
\newblock Deep residual learning for image recognition.
\newblock \emph{2016 IEEE Conference on Computer Vision and Pattern Recognition
  (CVPR)}, Jun 2016.
\newblock \doi{10.1109/cvpr.2016.90}.
\newblock URL \url{http://dx.doi.org/10.1109/CVPR.2016.90}.

\bibitem[Hochreiter et~al.(2001)Hochreiter, Younger, and
  Conwell]{hochreiter2001learning}
Sepp Hochreiter, A~Steven Younger, and Peter~R Conwell.
\newblock Learning to learn using gradient descent.
\newblock In \emph{International Conference on Artificial Neural Networks},
  pages 87--94. Springer, 2001.

\bibitem[Ioffe and Szegedy(2015)]{ioffe2015batch}
Sergey Ioffe and Christian Szegedy.
\newblock Batch normalization: Accelerating deep network training by reducing
  internal covariate shift.
\newblock \emph{arXiv preprint arXiv:1502.03167}, 2015.

\bibitem[Kahn et~al.(2017)Kahn, Villaflor, Pong, Abbeel, and
  Levine]{kahn2017uncertainty}
Gregory Kahn, Adam Villaflor, Vitchyr Pong, Pieter Abbeel, and Sergey Levine.
\newblock Uncertainty-aware reinforcement learning for collision avoidance.
\newblock \emph{arXiv preprint arXiv:1702.01182}, 2017.

\bibitem[Kingma and Ba(2014)]{kingma2014adam}
Diederik~P Kingma and Jimmy Ba.
\newblock Adam: A method for stochastic optimization.
\newblock \emph{arXiv preprint arXiv:1412.6980}, 2014.

\bibitem[Lakshminarayanan et~al.(2017)Lakshminarayanan, Pritzel, and
  Blundell]{lakshminarayanan2017simple}
Balaji Lakshminarayanan, Alexander Pritzel, and Charles Blundell.
\newblock Simple and scalable predictive uncertainty estimation using deep
  ensembles.
\newblock In \emph{Advances in Neural Information Processing Systems}, pages
  6402--6413, 2017.

\bibitem[Leike et~al.(2017)Leike, Martic, Krakovna, Ortega, Everitt, Lefrancq,
  Orseau, and Legg]{leike2017ai}
Jan Leike, Miljan Martic, Victoria Krakovna, Pedro~A Ortega, Tom Everitt,
  Andrew Lefrancq, Laurent Orseau, and Shane Legg.
\newblock Ai safety gridworlds.
\newblock \emph{arXiv preprint arXiv:1711.09883}, 2017.

\bibitem[Levine et~al.(2016)Levine, Finn, Darrell, and Abbeel]{levine2016end}
Sergey Levine, Chelsea Finn, Trevor Darrell, and Pieter Abbeel.
\newblock End-to-end training of deep visuomotor policies.
\newblock \emph{The Journal of Machine Learning Research}, 17\penalty0
  (1):\penalty0 1334--1373, 2016.

\bibitem[Li et~al.(2017)Li, Chen, Li, Gao, and Celikyilmaz]{li2017end}
Xiujun Li, Yun-Nung Chen, Lihong Li, Jianfeng Gao, and Asli Celikyilmaz.
\newblock End-to-end task-completion neural dialogue systems.
\newblock \emph{arXiv preprint arXiv:1703.01008}, 2017.

\bibitem[Lipton et~al.(2016)Lipton, Gao, Li, Chen, and
  Deng]{lipton2016combating}
Zachary~C Lipton, Jianfeng Gao, Lihong Li, Jianshu Chen, and Li~Deng.
\newblock Combating reinforcement learning’s sisyphean curse with intrinsic
  fear.(nov. 2016).
\newblock \emph{arXiv preprint cs.LG/1611.01211}, 2016.

\bibitem[Michelmore et~al.(2018)Michelmore, Kwiatkowska, and
  Gal]{michelmore2018evaluating}
Rhiannon Michelmore, Marta Kwiatkowska, and Yarin Gal.
\newblock Evaluating uncertainty quantification in end-to-end autonomous
  driving control, 2018.

\bibitem[Mnih et~al.(2015)Mnih, Kavukcuoglu, Silver, Rusu, Veness, Bellemare,
  Graves, Riedmiller, Fidjeland, Ostrovski, et~al.]{mnih2015human}
Volodymyr Mnih, Koray Kavukcuoglu, David Silver, Andrei~A Rusu, Joel Veness,
  Marc~G Bellemare, Alex Graves, Martin Riedmiller, Andreas~K Fidjeland, Georg
  Ostrovski, et~al.
\newblock Human-level control through deep reinforcement learning.
\newblock \emph{Nature}, 518\penalty0 (7540):\penalty0 529, 2015.

\bibitem[Mnih et~al.(2016)Mnih, Badia, Mirza, Graves, Lillicrap, Harley,
  Silver, and Kavukcuoglu]{mnih2016asynchronous}
Volodymyr Mnih, Adria~Puigdomenech Badia, Mehdi Mirza, Alex Graves, Timothy
  Lillicrap, Tim Harley, David Silver, and Koray Kavukcuoglu.
\newblock Asynchronous methods for deep reinforcement learning.
\newblock In \emph{International conference on machine learning}, pages
  1928--1937, 2016.

\bibitem[Paszke et~al.(2017)Paszke, Gross, Chintala, Chanan, Yang, DeVito, Lin,
  Desmaison, Antiga, and Lerer]{paszke2017automatic}
Adam Paszke, Sam Gross, Soumith Chintala, Gregory Chanan, Edward Yang, Zachary
  DeVito, Zeming Lin, Alban Desmaison, Luca Antiga, and Adam Lerer.
\newblock Automatic differentiation in {PyTorch}.
\newblock In \emph{NIPS Autodiff Workshop}, 2017.

\bibitem[Paul et~al.(2018)Paul, Osborne, and Whiteson]{paul2018fingerprint}
Supratik Paul, Michael~A Osborne, and Shimon Whiteson.
\newblock Fingerprint policy optimisation for robust reinforcement learning.
\newblock \emph{arXiv preprint arXiv:1805.10662}, 2018.

\bibitem[Saunders et~al.(2018)Saunders, Sastry, Stuhlmueller, and
  Evans]{saunders2018trial}
William Saunders, Girish Sastry, Andreas Stuhlmueller, and Owain Evans.
\newblock Trial without error: Towards safe reinforcement learning via human
  intervention.
\newblock In \emph{Proceedings of the 17th International Conference on
  Autonomous Agents and MultiAgent Systems}, pages 2067--2069. International
  Foundation for Autonomous Agents and Multiagent Systems, 2018.

\bibitem[Schmidhuber(1987)]{schmidhuber1987evolutionary}
J{\"u}rgen Schmidhuber.
\newblock \emph{Evolutionary principles in self-referential learning, or on
  learning how to learn: the meta-meta-... hook}.
\newblock PhD thesis, Technische Universit{\"a}t M{\"u}nchen, 1987.

\bibitem[Schulman et~al.(2017)Schulman, Wolski, Dhariwal, Radford, and
  Klimov]{schulman2017proximal}
John Schulman, Filip Wolski, Prafulla Dhariwal, Alec Radford, and Oleg Klimov.
\newblock Proximal policy optimization algorithms, 2017.

\bibitem[Silver et~al.(2016)Silver, Huang, Maddison, Guez, Sifre, Van
  Den~Driessche, Schrittwieser, Antonoglou, Panneershelvam, Lanctot,
  et~al.]{silver2016mastering}
David Silver, Aja Huang, Chris~J Maddison, Arthur Guez, Laurent Sifre, George
  Van Den~Driessche, Julian Schrittwieser, Ioannis Antonoglou, Veda
  Panneershelvam, Marc Lanctot, et~al.
\newblock Mastering the game of go with deep neural networks and tree search.
\newblock \emph{nature}, 529\penalty0 (7587):\penalty0 484, 2016.

\bibitem[Srivastava et~al.(2014)Srivastava, Hinton, Krizhevsky, Sutskever, and
  Salakhutdinov]{srivastava2014dropout}
Nitish Srivastava, Geoffrey Hinton, Alex Krizhevsky, Ilya Sutskever, and Ruslan
  Salakhutdinov.
\newblock Dropout: a simple way to prevent neural networks from overfitting.
\newblock \emph{The Journal of Machine Learning Research}, 15\penalty0
  (1):\penalty0 1929--1958, 2014.

\bibitem[Sutton and Barto(2018)]{sutton2018reinforcement}
Richard~S Sutton and Andrew~G Barto.
\newblock \emph{Reinforcement learning: An introduction}.
\newblock MIT press, 2018.

\bibitem[Thrun and Pratt(2012)]{thrun2012learning}
Sebastian Thrun and Lorien Pratt.
\newblock \emph{Learning to learn}.
\newblock Springer Science \& Business Media, 2012.

\bibitem[Tieleman and Hinton(2012)]{tieleman2012lecture}
Tijmen Tieleman and Geoffrey Hinton.
\newblock Lecture 6.5-rmsprop: Divide the gradient by a running average of its
  recent magnitude.
\newblock \emph{COURSERA: Neural networks for machine learning}, 4\penalty0
  (2):\penalty0 26--31, 2012.

\bibitem[Wang et~al.(2016)Wang, Kurth-Nelson, Tirumala, Soyer, Leibo, Munos,
  Blundell, Kumaran, and Botvinick]{wang2016learning}
Jane~X Wang, Zeb Kurth-Nelson, Dhruva Tirumala, Hubert Soyer, Joel~Z Leibo,
  Remi Munos, Charles Blundell, Dharshan Kumaran, and Matt Botvinick.
\newblock Learning to reinforcement learn.
\newblock \emph{arXiv preprint arXiv:1611.05763}, 2016.

\bibitem[Watkins and Dayan(1992)]{watkins1992q}
Christopher~JCH Watkins and Peter Dayan.
\newblock Q-learning.
\newblock \emph{Machine learning}, 8\penalty0 (3-4):\penalty0 279--292, 1992.

\bibitem[Zhang et~al.(2018)Zhang, Vinyals, Munos, and Bengio]{zhang2018study}
Chiyuan Zhang, Oriol Vinyals, Remi Munos, and Samy Bengio.
\newblock A study on overfitting in deep reinforcement learning.
\newblock \emph{arXiv preprint arXiv:1804.06893}, 2018.

\end{thebibliography}
\bibliographystyle{plainnat}

\appendix
\newpage
\section{Supplementary Material} \label{sec:Appendix}

\begin{figure}[ht]
  \centering
  \begin{subfigure}[l]{0.15\linewidth}
    \includegraphics[width=\linewidth]{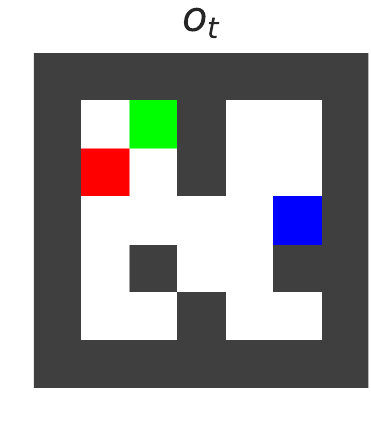}
  \end{subfigure}
  ~
  \begin{subfigure}[l]{0.15\linewidth}
    \includegraphics[width=\linewidth]{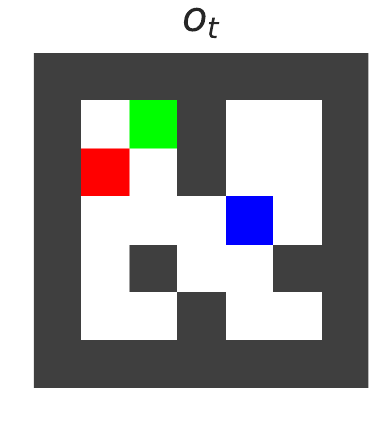}
  \end{subfigure}
  ~
  \begin{subfigure}[l]{0.15\linewidth}
    \includegraphics[width=\linewidth]{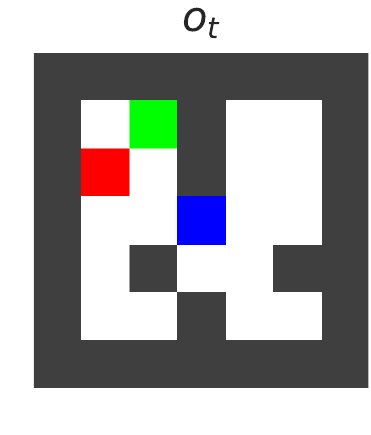}
  \end{subfigure}
  ~
  \begin{subfigure}[l]{0.15\linewidth}
    \includegraphics[width=\linewidth]{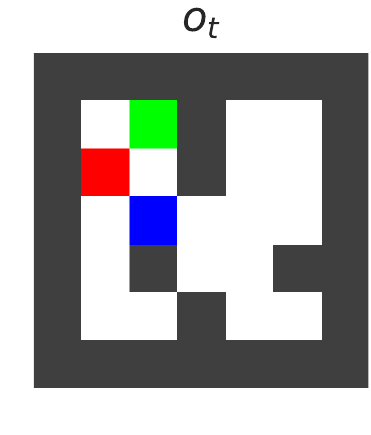}
  \end{subfigure}
  ~
  \begin{subfigure}[l]{0.15\linewidth}
    \includegraphics[width=\linewidth]{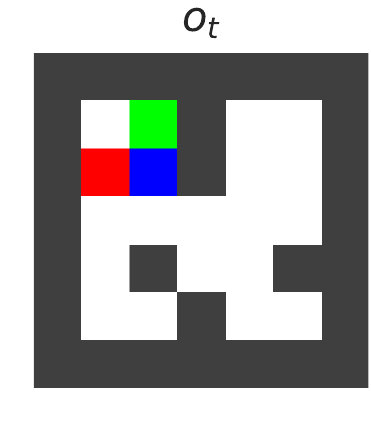}
  \end{subfigure}
  ~
  \begin{subfigure}[l]{0.15\linewidth}
    \includegraphics[width=\linewidth]{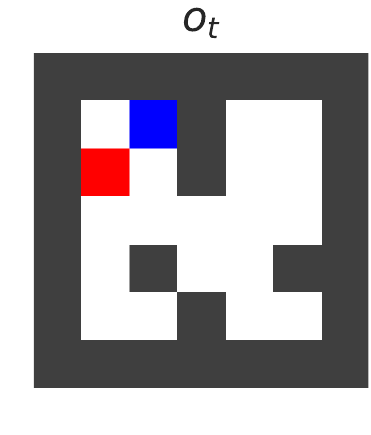}
  \end{subfigure}
  \caption{Example trajectory from a \texttt{Full} environment. Agent: blue. Goal: green. Lava: red. Walls: grey.}
  \label{fig:full_obs}
\end{figure}

\paragraph{Full Setting}
See Fig.\ref{fig:full_obs} for some frames from the \texttt{Full} gridworld setting.

\begin{figure}[h]

  \begin{subfigure}[b]{\linewidth}
    \centering
    \includegraphics[width=0.25\linewidth]{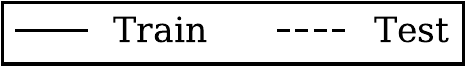}
  \end{subfigure}

  \begin{subfigure}[b]{\linewidth}
    \includegraphics[width=\linewidth]{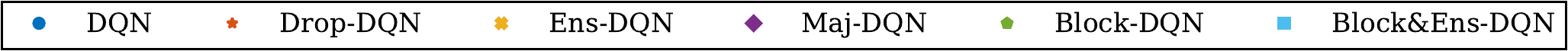}
  \end{subfigure}

  \begin{subfigure}[t]{0.48\linewidth}
    \includegraphics[width=\linewidth]{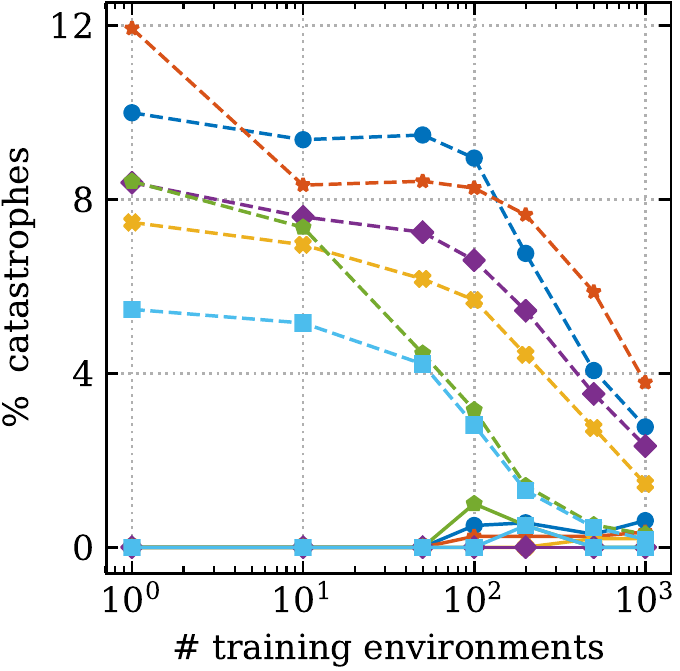}
    \caption{Percentage of catastrophic outcomes (lower is better),
    as a function of number of training environments.}
    \label{fig:full::catastrophes_complete}
  \end{subfigure}
  ~
  \begin{subfigure}[t]{0.48\linewidth}
    \includegraphics[width=\linewidth]{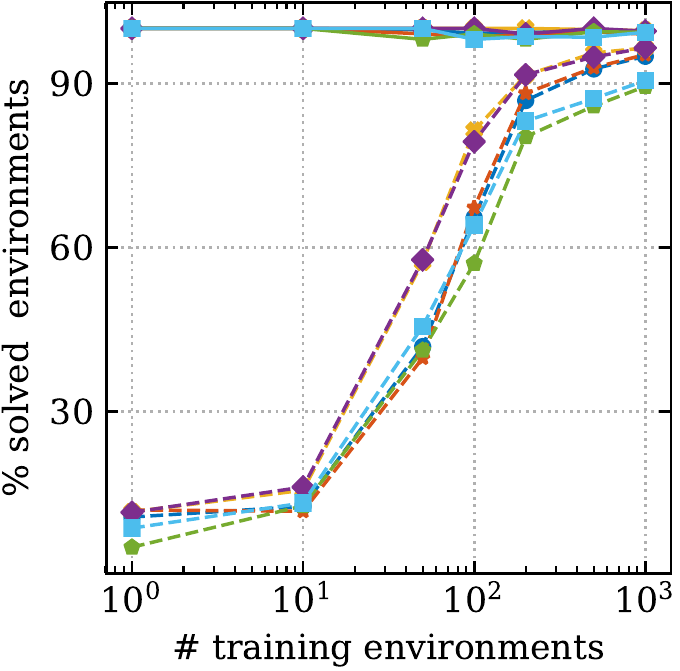}
    \caption{Percentage of solved environments (higher is better),
    as a function of number of training environments.}
    \label{fig:full::solved_complete}
  \end{subfigure}

  \caption{Complete quantitative experimental results on the \texttt{Full} setting,
  trained to convergence. Nine seeds are used for training the
  agents and the mean performances are visualized.}
  \label{fig:full::quant_results_complete}
\end{figure}

\begin{figure}[h]

  \begin{subfigure}[b]{\linewidth}
    \centering
    \includegraphics[width=0.25\linewidth]{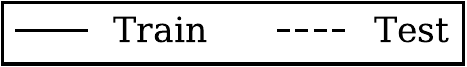}
  \end{subfigure}

  \begin{subfigure}[b]{\linewidth}
    \includegraphics[width=\linewidth]{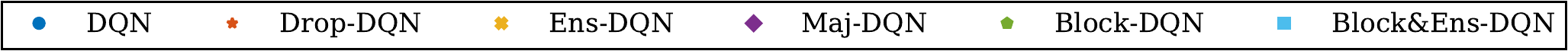}
  \end{subfigure}

  \begin{subfigure}[t]{0.48\linewidth}
    \includegraphics[width=\linewidth]{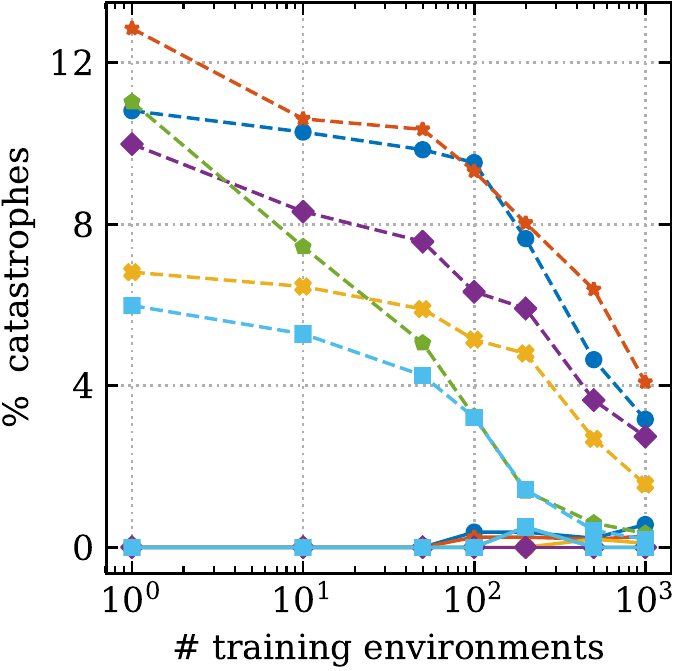}
    \caption{Percentage of catastrophic outcomes (lower is better),
    as a function of number of training environments.}
    \label{fig:reveal::catastrophes_complete}
  \end{subfigure}
  ~
  \begin{subfigure}[t]{0.48\linewidth}
    \includegraphics[width=\linewidth]{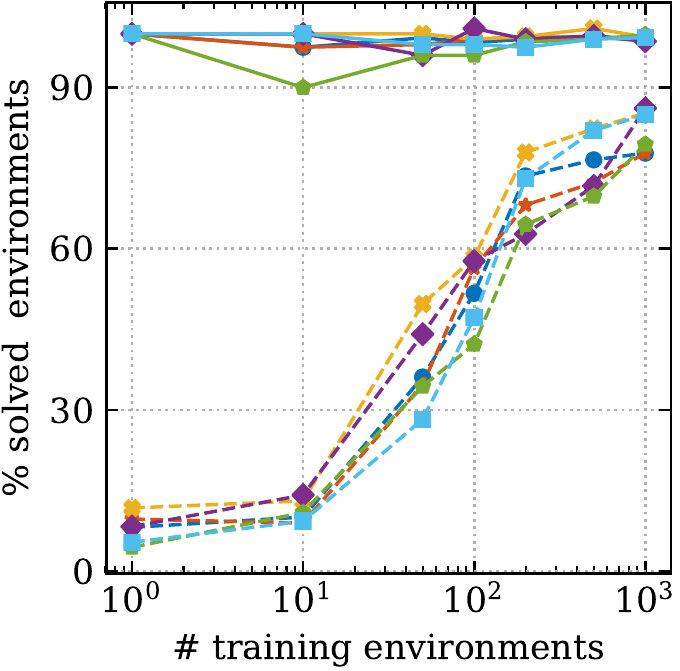}
    \caption{Percentage of solved environments (higher is better),
    as a function of number of training environments.}
    \label{fig:reveal::solved_complete}
  \end{subfigure}

  \caption{Complete quantitative experimental results on the \texttt{Reveal} setting,
  trained to convergence. Nine seeds are used for training the
  agents and the mean performances are visualized.}
  \label{fig:reveal::quant_results_complete}
\end{figure}

\paragraph{Further Results.}
Shown in Fig.~\ref{fig:full::quant_results_complete} are the results for all the methods on the \texttt{Full} setting. See Fig.~\ref{fig:reveal::quant_results_complete} for results on the \texttt{Reveal} setting. Shown are also the performance on the training environments (solid lines). We see similar results to the main paper, and note that as expected the \texttt{Full} setting has better generalization performance for \% solved than \texttt{Reveal}, but the catastrophe performance is similar in each.

\begin{figure}[h]

  \begin{subfigure}[b]{\linewidth}
    \centering
    \includegraphics[width=0.25\linewidth]{fig/experiments/methods/reveal/legend2}
  \end{subfigure}

  \begin{subfigure}[b]{\linewidth}
    \includegraphics[width=\linewidth]{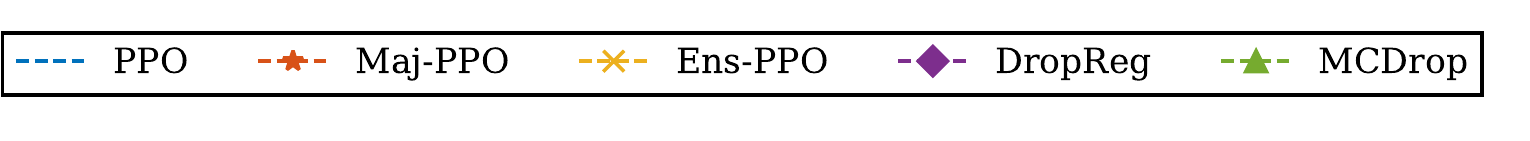}
  \end{subfigure}

  \begin{subfigure}[t]{0.48\linewidth}
    \includegraphics[width=\linewidth]{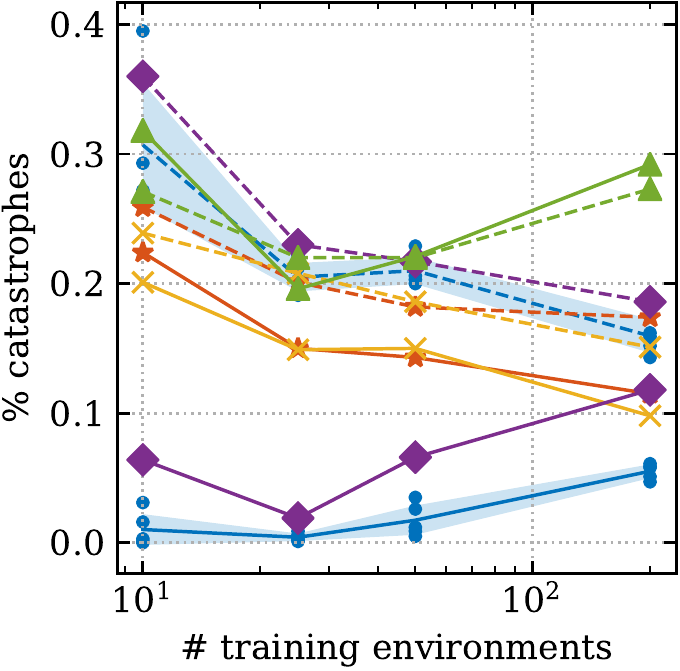}
    \caption{Percentage of catastrophic outcomes (lower is better),
    as a function of number of training environments.}
    \label{fig:coinrun::catastrophes_complete}
  \end{subfigure}
  ~
  \begin{subfigure}[t]{0.48\linewidth}
    \includegraphics[width=\linewidth]{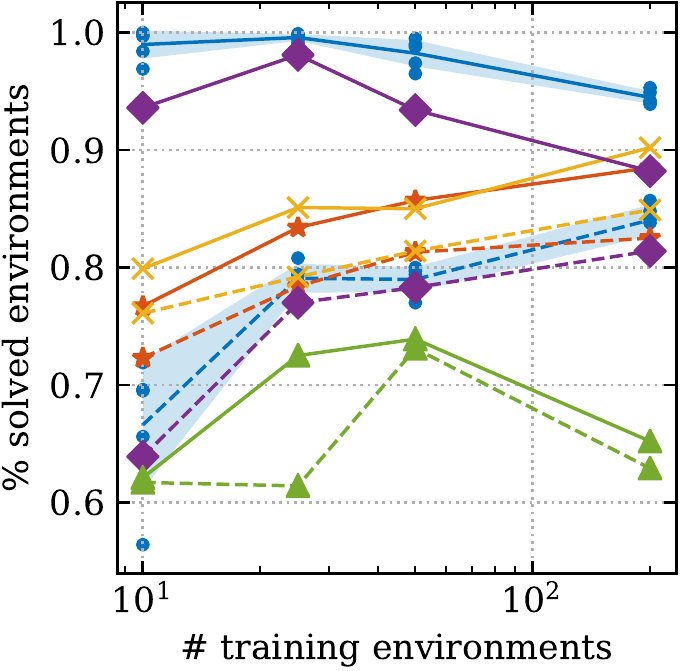}
    \caption{Percentage of solved environments (higher is better),
    as a function of number of training environments.}
    \label{fig:coinrun::solved_complete}
  \end{subfigure}

  \caption{Complete quantitative experimental results on the CoinRun setting.}
  \label{fig:coinrun::quant_results_complete}
\end{figure}

See Fig.~\ref{fig:coinrun::quant_results_complete} for complete results on CoinRun, including training performance and dropout as both a regularizer (turned off at test) or for MC dropout (dropout on at test time). 

\begin{figure}[ht]
    
  \begin{subfigure}[b]{\linewidth}
    \centering
        \includegraphics[width=\linewidth]{fig/experiments/coinrun/rocs/main_legend.pdf}
  \end{subfigure}

  \begin{subfigure}[t]{0.23\linewidth}
  \centering
    \includegraphics[width=\linewidth]{fig/experiments/coinrun/rocs/Nlev_10Train_False_dt_1.png}
    \caption{10 levs, $dt=1$}
    \label{fig:app_app_roc:10t1}
  \end{subfigure}
  ~
  \begin{subfigure}[t]{0.23\linewidth}
  \centering
    \includegraphics[width=\linewidth]{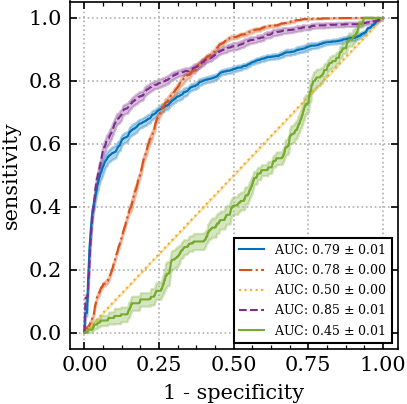}
    \caption{10 levs, $dt=3$}
    \label{fig:app_app_roc:10t10}
  \end{subfigure}
~
  \begin{subfigure}[t]{0.23\linewidth}
  \centering
    \includegraphics[width=\linewidth]{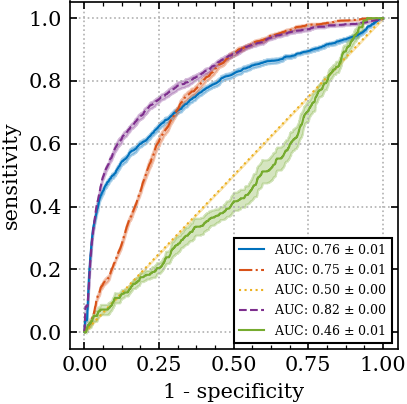}
    \caption{10 levs, $dt=5$}
    \label{fig:app_app_roc:25t1}
  \end{subfigure}
    ~
  \begin{subfigure}[t]{0.23\linewidth}
  \centering
    \includegraphics[width=\linewidth]{fig/experiments/coinrun/rocs/Nlev_10Train_False_dt_10.png}
    \caption{10 levs, $dt=10$}
    \label{fig:app_app_roc:25t10}
  \end{subfigure}
  
    \vspace{1em}
    
     \begin{subfigure}[t]{0.23\linewidth}
  \centering
    \includegraphics[width=\linewidth]{fig/experiments/coinrun/rocs/Nlev_25Train_False_dt_1.png}
    \caption{25 levs, $dt=1$}
    \label{fig:app_app_roc:10t1}
  \end{subfigure}
  ~
  \begin{subfigure}[t]{0.23\linewidth}
  \centering
    \includegraphics[width=\linewidth]{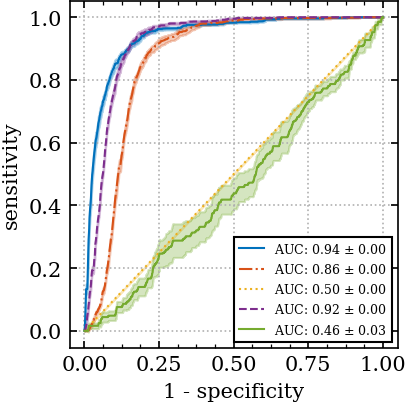}
    \caption{25 levs, $dt=3$}
    \label{fig:app_app_roc:10t10}
  \end{subfigure}
~
  \begin{subfigure}[t]{0.23\linewidth}
  \centering
    \includegraphics[width=\linewidth]{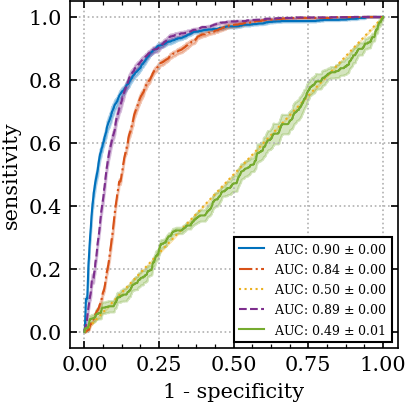}
    \caption{25 levs, $dt=5$}
    \label{fig:app_app_roc:25t1}
  \end{subfigure}
    ~
  \begin{subfigure}[t]{0.23\linewidth}
  \centering
    \includegraphics[width=\linewidth]{fig/experiments/coinrun/rocs/Nlev_25Train_False_dt_10.png}
    \caption{25 levs, $dt=10$}
    \label{fig:app_app_roc:25t10}
  \end{subfigure} 
  
      \vspace{1em}
    
     \begin{subfigure}[t]{0.23\linewidth}
  \centering
    \includegraphics[width=\linewidth]{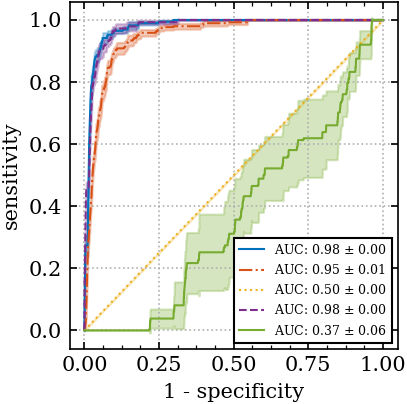}
    \caption{50 levs, $dt=1$}
    \label{fig:app_app_roc:10t1}
  \end{subfigure}
  ~~
  \begin{subfigure}[t]{0.23\linewidth}
  \centering
    \includegraphics[width=\linewidth]{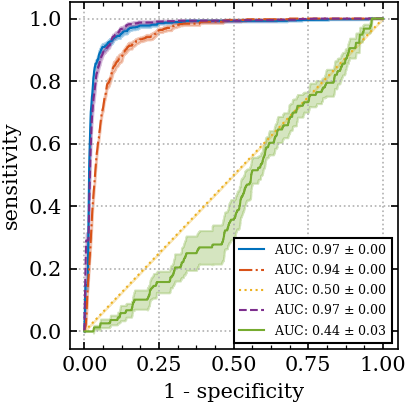}
    \caption{50 levs, $dt=3$}
    \label{fig:app_app_roc:10t10}
  \end{subfigure}
~
  \begin{subfigure}[t]{0.23\linewidth}
  \centering
    \includegraphics[width=\linewidth]{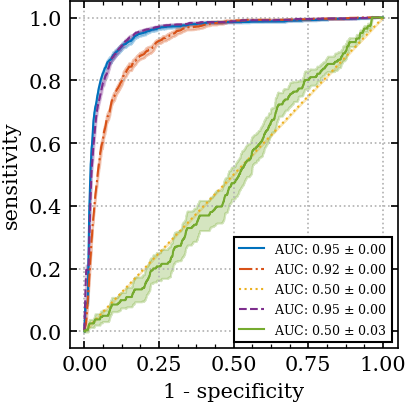}
    \caption{50 levs, $dt=5$}
    \label{fig:app_app_roc:25t1}
  \end{subfigure}
    ~
  \begin{subfigure}[t]{0.23\linewidth}
  \centering
    \includegraphics[width=\linewidth]{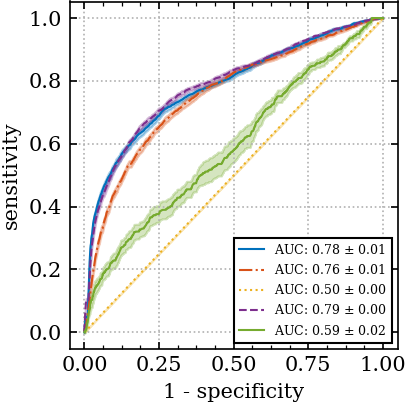}
    \caption{50 levs, $dt=10$}
    \label{fig:app_app_roc:25t10}
  \end{subfigure} 
  
      \vspace{1em}
    
     \begin{subfigure}[t]{0.23\linewidth}
  \centering
    \includegraphics[width=\linewidth]{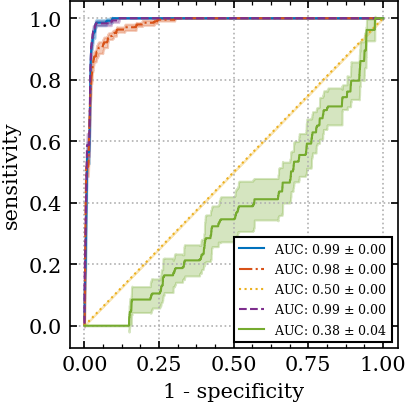}
    \caption{200 levs, $dt=1$}
    \label{fig:app_app_roc:10t1}
  \end{subfigure}
  ~
  \begin{subfigure}[t]{0.23\linewidth}
  \centering
    \includegraphics[width=\linewidth]{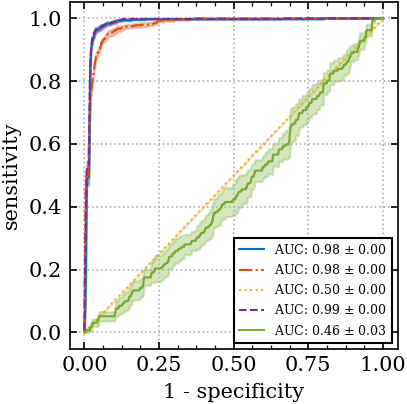}
    \caption{200 levs, $dt=3$}
    \label{fig:app_app_roc:10t10}
  \end{subfigure}
~
  \begin{subfigure}[t]{0.23\linewidth}
  \centering
    \includegraphics[width=\linewidth]{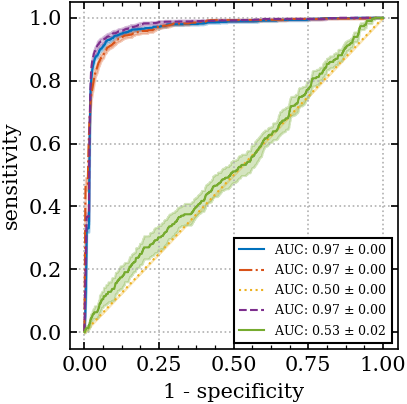}
    \caption{200 levs, $dt=5$}
    \label{fig:app_app_roc:25t1}
  \end{subfigure}
    ~
  \begin{subfigure}[t]{0.23\linewidth}
  \centering
    \includegraphics[width=\linewidth]{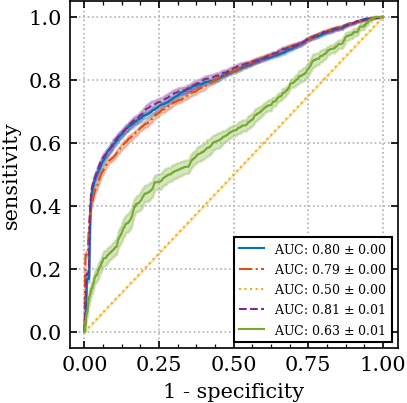}
    \caption{200 levs, $dt=10$}
    \label{fig:app_app_roc:25t10}
  \end{subfigure} 

  \caption{ROC Curves for binary classifier based on discrimination function $U = \alpha \mu + \beta \sigma$, a combination of mean and standard deviation of the agents' value functions, classifying a catastrophe occurring in the next $dt$ steps.  All plots are on data from one episode on 1000 different test levels. 
  Towards the top left is better.
  Higher AUC is better.
  \textbf{Columns:}  $dt=1,3,5,10$. 
  \textbf{Rows:} $10,25,50,200$ train levels. }
  \label{fig:app_roc:test}
\end{figure}

\begin{figure}[ht]
    
  \begin{subfigure}[b]{\linewidth}
    \centering
        \includegraphics[width=\linewidth]{fig/experiments/coinrun/rocs/main_legend.pdf}
  \end{subfigure}

  \begin{subfigure}[t]{0.23\linewidth}
  \centering
    \includegraphics[width=\linewidth]{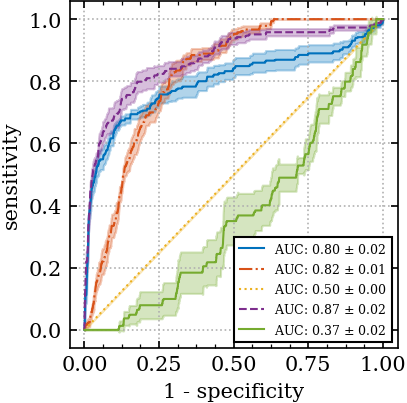}
    \caption{10 levs, $dt=1$}
    \label{fig:app_app_roc:10t1}
  \end{subfigure}
  ~
  \begin{subfigure}[t]{0.23\linewidth}
  \centering
    \includegraphics[width=\linewidth]{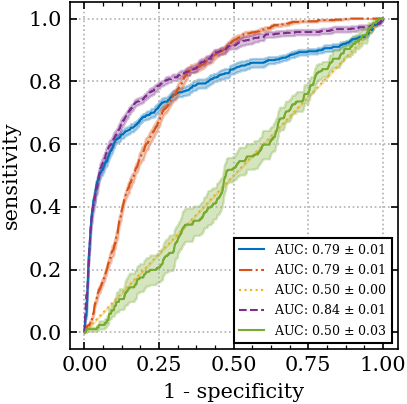}
    \caption{10 levs, $dt=3$}
    \label{fig:app_app_roc:10t10}
  \end{subfigure}
~
  \begin{subfigure}[t]{0.23\linewidth}
  \centering
    \includegraphics[width=\linewidth]{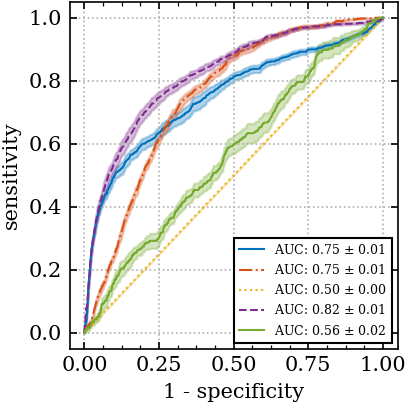}
    \caption{10 levs, $dt=5$}
    \label{fig:app_app_roc:25t1}
  \end{subfigure}
    ~
  \begin{subfigure}[t]{0.23\linewidth}
  \centering
    \includegraphics[width=\linewidth]{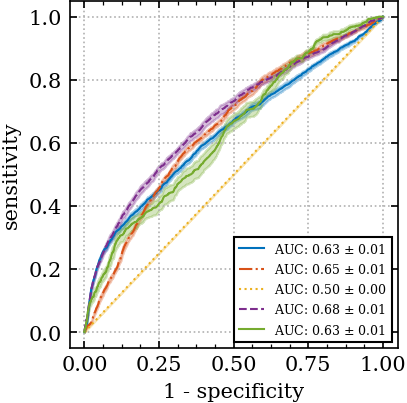}
    \caption{10 levs, $dt=10$}
    \label{fig:app_app_roc:25t10}
  \end{subfigure}
  
    \vspace{1em}
    
     \begin{subfigure}[t]{0.23\linewidth}
  \centering
    \includegraphics[width=\linewidth]{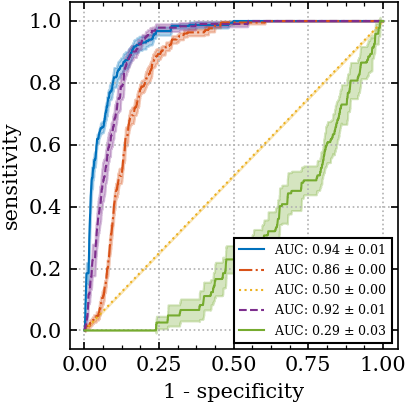}
    \caption{25 levs, $dt=1$}
    \label{fig:app_app_roc:10t1}
  \end{subfigure}
  ~
  \begin{subfigure}[t]{0.23\linewidth}
  \centering
    \includegraphics[width=\linewidth]{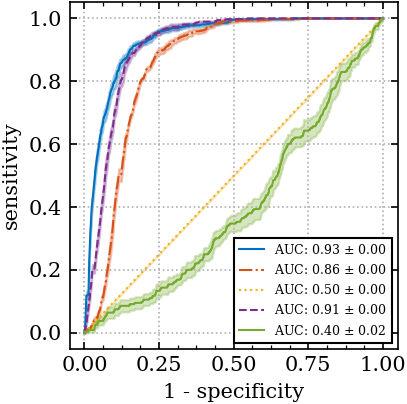}
    \caption{25 levs, $dt=3$}
    \label{fig:app_app_roc:10t10}
  \end{subfigure}
~
  \begin{subfigure}[t]{0.23\linewidth}
  \centering
    \includegraphics[width=\linewidth]{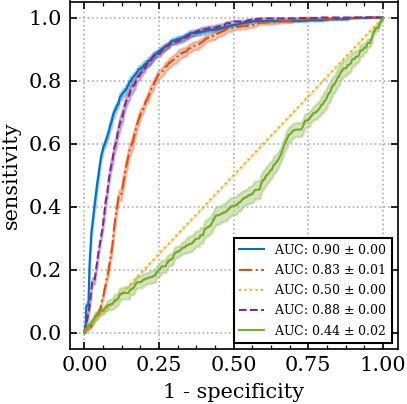}
    \caption{25 levs, $dt=5$}
    \label{fig:app_app_roc:25t1}
  \end{subfigure}
    ~
  \begin{subfigure}[t]{0.23\linewidth}
  \centering
    \includegraphics[width=\linewidth]{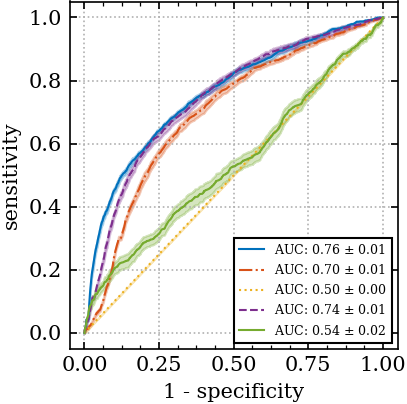}
    \caption{25 levs, $dt=10$}
    \label{fig:app_app_roc:25t10}
  \end{subfigure} 
  
      \vspace{1em}
    
     \begin{subfigure}[t]{0.23\linewidth}
  \centering
    \includegraphics[width=\linewidth]{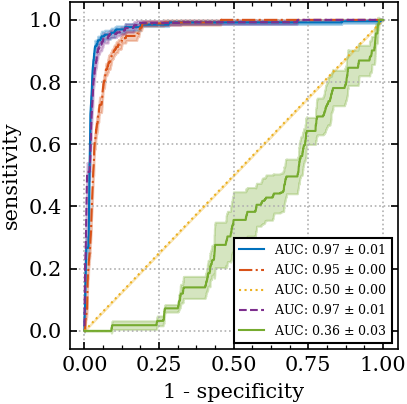}
    \caption{50 levs, $dt=1$}
    \label{fig:app_app_roc:10t1}
  \end{subfigure}
  ~~
  \begin{subfigure}[t]{0.23\linewidth}
  \centering
    \includegraphics[width=\linewidth]{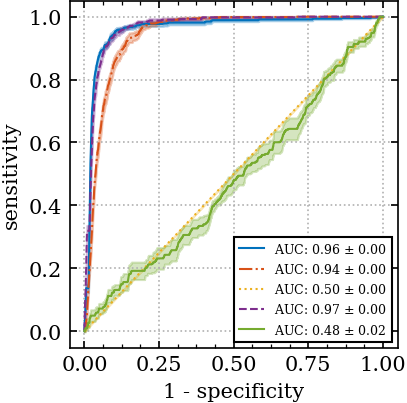}
    \caption{50 levs, $dt=3$}
    \label{fig:app_app_roc:10t10}
  \end{subfigure}
~
  \begin{subfigure}[t]{0.23\linewidth}
  \centering
    \includegraphics[width=\linewidth]{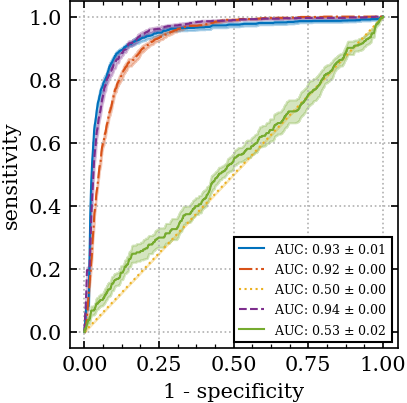}
    \caption{50 levs, $dt=5$}
    \label{fig:app_app_roc:25t1}
  \end{subfigure}
    ~
  \begin{subfigure}[t]{0.23\linewidth}
  \centering
    \includegraphics[width=\linewidth]{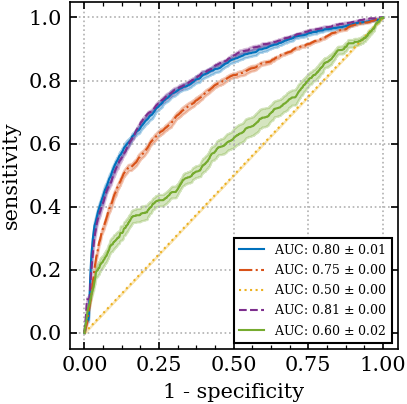}
    \caption{50 levs, $dt=10$}
    \label{fig:app_app_roc:25t10}
  \end{subfigure} 
  
      \vspace{1em}
    
     \begin{subfigure}[t]{0.23\linewidth}
  \centering
    \includegraphics[width=\linewidth]{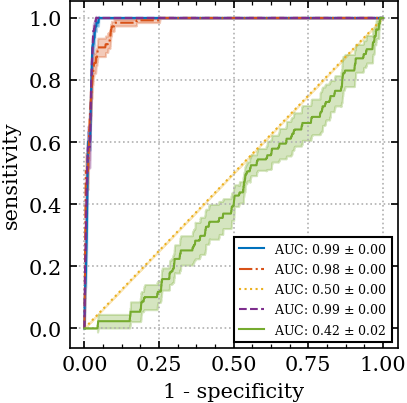}
    \caption{200 levs, $dt=1$}
    \label{fig:app_app_roc:10t1}
  \end{subfigure}
  ~
  \begin{subfigure}[t]{0.23\linewidth}
  \centering
    \includegraphics[width=\linewidth]{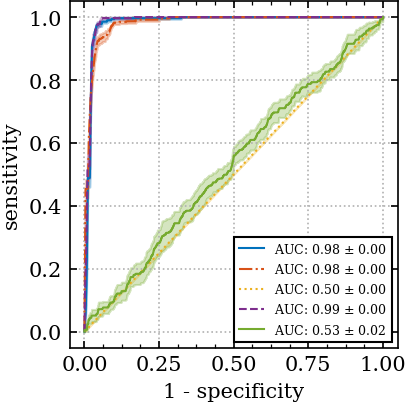}
    \caption{200 levs, $dt=3$}
    \label{fig:app_app_roc:10t10}
  \end{subfigure}
~
  \begin{subfigure}[t]{0.23\linewidth}
  \centering
    \includegraphics[width=\linewidth]{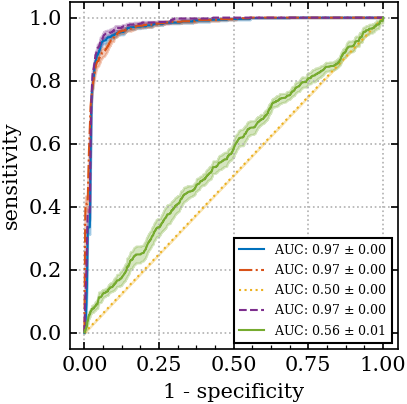}
    \caption{200 levs, $dt=5$}
    \label{fig:app_app_roc:25t1}
  \end{subfigure}
    ~
  \begin{subfigure}[t]{0.23\linewidth}
  \centering
    \includegraphics[width=\linewidth]{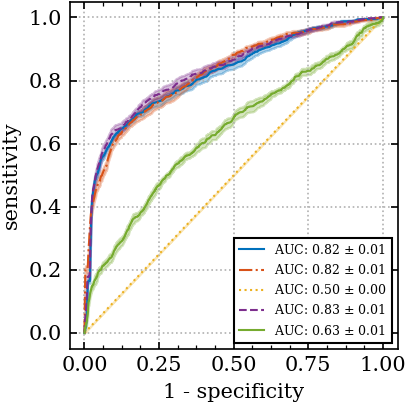}
    \caption{200 levs, $dt=10$}
    \label{fig:app_app_roc:25t10}
  \end{subfigure} 

  \caption{ROC Curves evaluated on training levels.
  \textbf{Columns:}  $dt=1,3,5,10$. 
  \textbf{Rows:} $10,25,50,200$ train levels. }
  \label{fig:app_roc:train}
\end{figure}

See Fig.~\ref{fig:app_roc:test} and Fig.~\ref{fig:app_roc:train} for ROC curves on test and train environments respectively, for $dt=1,3,5,10$ and $10,25,50,200$ training levels.

\subsection{Algorithm Settings}\label{sec:app:algo}
\paragraph{PPO settings}
We trained our agents each for $T=1e7$ timesteps, using a linearly decaying learning rate (initial value $2e-4$) and Adam optimizer \citep{kingma2014adam}.
We used 256 PPO steps, 8 minibatches, 3 PPO epochs, entropy coefficient of 0.01, and a decay rate of 0.999.
We used the same IMPALA-CNN style architecture as \citet{cobbe2018quantifying} (itself taken from \cite{espeholt2018impala}), except we modify it to be smaller, using a convolutional layer with 5 filters, max pooling with pool size 3 and strides 2 and same padding, followed by two residual blocks, each containing [relu, conv, relu, conv] to which the input is added to the output in residual style \cite{He_2016}. This is followed by a relu and fully connected layer to 256 hidden units, followed by another fully connected layer with 6 heads, one for each action logit in a categorical distribution. 
For the dropout agent we decayed the dropout probability according to $\max[0.01, d]$, where $d$ decays linearly from 0.1 to zero. 

Training was performed using 32-CPU machines, using TensorFlow \cite{tensorflow2015-whitepaper} for PPO and PyTorch \cite{paszke2017automatic} for DQN. 

\end{document}